\documentclass[acmlarge]{acmart}
\makeatletter
\newcommand{\confshort}{\acmConference@shortname}
\newcommand{\conffull}{\acmConference@name}
\newcommand{\confdate}{\acmConference@date}
\newcommand{\confloc}{\acmConference@venue}
\AtBeginDocument{
  \fancypagestyle{firstpagestyle}{
    \fancyhead{}%
    \fancyfoot[C]{}%
  }
  \fancyhf{}
  \fancyhead[LO]{\@headfootfont\shorttitle}%
  \fancyhead[RE]{\@headfootfont\@shortauthors}%
  \fancyhead[LE]{\@headfootfont\footnotesize \confshort, \confdate, \confloc}%
  \fancyhead[RO]{\@headfootfont\footnotesize \confshort, \confdate, \confloc}%
  \fancyfoot[C]{}%
}
\makeatother
\acmBooktitle{\conffull\@ (\confshort), \confdate, \confloc}

\AtBeginDocument{%
  }

\setcopyright{acmlicensed}
\copyrightyear{2026}
\acmYear{2026}
\acmDOI{XXXXXXX.XXXXXXX}
\acmConference[FAccT '26]{The 2026 ACM Conference on Fairness, Accountability, and Transparency}{June 25--28, 2026}{Montreal, Canada}

\acmISBN{978-1-4503-XXXX-X/2018/06}

\usepackage{caption}
\usepackage{hyperref}
\usepackage{url}
\usepackage{graphicx}
\usepackage{multirow}%
\usepackage{float}
\usepackage{placeins}
\usepackage{pdflscape}
\usepackage{pifont}
\newtheorem{definition}{Definition}%
\usepackage[table]{xcolor}
\usepackage{geometry}
\usepackage{graphicx}
\usepackage{multirow}
\usepackage{rotating} 

\definecolor{lightgray}{gray}{0.92}
\definecolor{darkgreen}{rgb}{0,0.5,0}

\copyrightyear{2026}
\acmYear{2026}
\setcopyright{cc}
\setcctype{by}
\acmConference[FAccT '26]{The 2026 ACM Conference on Fairness, Accountability, and Transparency}{June 25--28, 2026}{Montreal, QC, Canada}
\acmBooktitle{The 2026 ACM Conference on Fairness, Accountability, and Transparency (FAccT '26), June 25--28, 2026, Montreal, QC, Canada}
\acmDOI{10.1145/3805689.3812361}
\acmISBN{979-8-4007-2596-8/2026/06}

\begin{document}

\title{Quantifying the Privacy of Counterfactuals by Leveraging Membership Inference Attacks Against Synthetic Data}

\author{Maryam Babaei}
\email{maryam.babaei.1@ens.etsmtl.ca}
\orcid{0009-0006-2071-7123}
\affiliation{%
  \institution{ÉTS Montreal and Mila}
  \country{Canada}
}

\author{Yingke Wang}
\affiliation{%
  \institution{ÉTS Montreal and Mila}
  \country{Canada}
  }
\email{yingke.wang@etsmtl.ca}

\author{Hadrien Lautraite}
\affiliation{%
  \institution{UQAM}
  \country{Canada}
}
\email{lautraite.hadrien@courrier.uqam.ca}

\author{Héber H. Arcolezi}
\affiliation{%
 \institution{ÉTS Montreal}
 \country{Canada}}
 \affiliation{%
 \institution{Inria Grenoble}
 \country{France}}
 \email{heber.hwang-arcolezi@etsmtl.ca}

\author{Ulrich Aïvodji}
\affiliation{%
  \institution{ÉTS Montreal and Mila}
  \country{Canada}
  }
\email{ulrich.aivodji@etsmtl.ca}
  
\author{Sébastien Gambs}
\affiliation{%
  \institution{UQAM}
  \country{Canada}}
\email{gambs.sebastien@uqam.ca}

\renewcommand{\shortauthors}{Babaei et al.}
\renewcommand{\shorttitle}{Quantifying the Privacy of CFs by Leveraging MIA Against Synthetic Data}
\begin{abstract}
Counterfactuals are typically used in high-stakes decision areas to explain a machine learning model by showing how changes to the user profiles result in the desired outcome.
However, explaining the model's decisions through counterfactuals can also be exploited by an adversary to conduct privacy attacks against the model or its training data.
Drawing on the analogy that counterfactuals provide realistic substitutes for real training data, similar to synthetic data, we demonstrate in this paper how it is possible to successfully perform privacy attacks on counterfactuals by drawing on the attacks developed against synthetic data. 
More precisely, we investigate the effectiveness of the membership inference attacks designed for synthetic data on various types of counterfactuals.
Additionally, while existing membership inference attacks against counterfactuals usually require to be able to query the model, we show how it is possible to perform successful membership inference attacks using only a set of counterfactuals, with no access to the model from which they are generated.
Our results demonstrate that model developers should be more cautious when releasing counterfactuals to various users, as it can lead to a privacy breach.
\end{abstract}



\maketitle

\section{Introduction}
\label{chap:introduction}

Counterfactuals are instances generated to show the most similar profiles to the query profile that achieve the desired outcome~\cite{wachter2017counterfactual,mothilal2020explaining,karimi2022survey,brughmans2023nice}.
In particular, they are often used in machine learning (ML) for high-stakes decision settings to help users understand the model's decisions. 
However, counterfactuals can also reveal information about the model itself or the training data through privacy attacks~\cite{aivodji2020model,kuppa2021adversarial,wang2022dualcf,goethals2022privacy} such as membership inference attacks (MIAs)~\cite{shokri2017membership}.
Yet, few MIA attacks have been developed against counterfactuals~\cite{pawelczyk2023privacy}, and additionally, they need query access to the model.
Thus, model providers can potentially prevent such attacks by limiting the number of queries per user or providing repetitive counterfactuals for similar queries. 
In this paper, we investigate how an adversary could circumvent such protection mechanisms by leveraging MIAs designed against synthetic data in the so-called \emph{no-box setting}, in which only the set of counterfactuals generated by the model is available to the adversary.  
As synthetic data is often used for sharing purposes when the privacy of training data is critical, a wide range of research has been conducted to evaluate these data against privacy attacks such as membership inference attacks~\cite{hilprecht2019monte,chen2020gan,van2023membership}.
Furthermore, MIAs developed to target synthetic data often do not require access to the models generating this data~\cite{owen2013monte,chen2020gan}.
Rather, they are designed to predict the membership of a target instance using only the synthetic data generated, which is consistent with the definition of a no-box attack setting~\cite{chen2017zoo}. 
However, some of these attacks also assume the availability of auxiliary data drawn from the same distribution as the training set~\cite{van2023membership}.

Since counterfactuals and synthetic data both try to generate instances similar to the training set, in this paper, we propose to view counterfactuals as artifacts produced by the counterfactual generation process, in the same manner as synthetic data can be considered as being derived from the training data through a generative process.
While MIAs against synthetic data have been deeply explored, their applicability to counterfactuals remains a critical gap in the literature. 
Considering counterfactuals' similarity to synthetic data, the main contribution of this work is to bridge these two domains by investigating the transferability of state-of-the-art synthetic data attacks to the counterfactual landscape.
More precisely, we aim at determining when releasing counterfactuals is safe when limiting queries per user and whether it is possible to perform no-box attacks against them, i.e., a significant shift from existing literature, which often assumes stronger adversary capabilities.
More precisely, we have implemented an ensembling MIA~\cite{ward2025ensembling} against counterfactuals generated by state-of-the-art counterfactual generation mechanisms~\cite{wachter2017counterfactual,brughmans2023nice,mothilal2020explaining}
and compared their effectiveness with that of counterfactual distance attack, an MIA attack designed specifically for counterfactuals~\cite{pawelczyk2023privacy}.
Our approach has the additional benefit of working in the no-box setting, which is a weaker adversary model than the one usually considered for attacks against counterfactuals.

The outline of the paper is as follows.
First, in Section~\ref{sec:mia_synthetic data}, we review the background on synthetic data and MIAs against such data before presenting in Section~\ref{sec:cfgeneration} the counterfactual generation mechanisms that we consider, as well as the counterfactual distance attack against which we will compare ourselves. 
Afterwards, in Section~\ref{chap:methodology}, we explain our attack framework and methodology, followed by the reporting of our experimental results in Section~\ref{chap:experimental-results}. 
Finally, we conclude in Section~\ref{chap:conclusion}.


\section{Membership inference attacks against synthetic data} 
\label{sec:mia_synthetic data}

Synthetic data generation approaches aim to generate realistic data that mimics the characteristics of training data.
For instance, synthetic data can be obtained by using generative models such as GANs (\emph{Generative adversarial networks})~\cite{goodfellow2014generative,bauer2024comprehensive}.
More precisely, the synthetic data should be diverse, novel and realistic with respect to the properties of the original data distribution~\cite{nikolenko2021synthetic}.
Since privacy is a critical concern in generating synthetic data, especially in domains such as health and finance in which the training data is highly personal, synthetic data generation methods have also been proposed based on the use of differential privacy~\cite{dwork2014algorithmic}. 
These include DP-synthetic data generation techniques that are GAN-based~\cite{beaulieu2019privacy,vietri2022private} or marginal-based~\cite{mckenna2021winning,mckenna2022aim}.

One of the popular privacy attacks against synthetic data is MIA, which aims to determine whether a specific profile was used during the synthetic data generator's training phase. 
This attack, which has been introduced by Hayes and collaborators~\cite{hayes2017logan}, can be formalized for synthetic data generators as follows~\cite{van2023membership}:
\begin{definition}
Let the random variable $X$ be defined on $\mathcal{X}$, with distribution $P_R(X)$. Let $D_{mem} \overset{\text{iid}}{\sim} P_R(X)$ be a training set of independently sampled points from distribution $P_R(X)$. 
Now let $G: Z \rightarrow X$ be a generator that generates data given some random (\emph{e.g.}, Gaussian) noise $Z$.
The generator $G$ is trained on $D_{mem}$ and is subsequently used to generate a synthetic dataset $D_{syn}$. 
Finally, let $A: X \rightarrow [0, 1]$ be the attacker model that possesses the synthetic dataset $D_{syn}$, some test point $x^*$, with $X^* \sim P_R(X)$, and possibly other knowledge (i.e., a reference dataset independently sampled from $P_R(X)$).
The adversary $A$ aims to determine whether $X^* \sim P_R(X)$ belongs to $D_{mem}$, hence the perfect attacker outputs $A({x^*}) = {1}[{x^*} \in D_{mem}]$.
\end{definition}
The main adversarial models used in this attack setting for MIA in synthetic data are the following:
\begin{itemize}
    \item \emph{Black-box setting}, in which the adversary can only blindly collect samples. 
    They may also have access to a reference dataset independently sampled from the training data distribution. 
    \item The \emph{white-box setting} in which the adversary has access to the model generating synthetic data and its internal settings, in addition to generated samples.
    \item The \emph{partial white-box setting}, in which the adversary has some level of access to the model and some information about the training dataset. 
\end{itemize}
In this paper, we performed the ensembling MIA in which six main state-of-the-art MIAs have been implemented.
These attacks are detailed hereafter. 

\textbf{\emph{Distance to Closest Record (DCR/ DCR-Diff)}}~\cite{chen2020gan}. Different variants of this attack exist in the black-box, partial white-box and white-box settings.
In the black-box setting, in which the adversary can only blindly collect samples generated by the generator, these samples are used to estimate the probability of one instance being a member of the training data of the generative model. 
More precisely, if the synthetic samples are closer to the target instance than other points from the same distribution, it is inferred as a member.
Equation~\ref{formula:ganleak} shows how the membership probability of $x$ is computed, in which $\phi(.,.)$ is the kernel function and $L(.,.)$ is the distance metric used. 
\begin{equation}
    P(m_i |x_i,\theta_v) \propto P_{\mathcal{G}_v} (x|\theta_v)\mathrm{.}  
\end{equation}

\begin{equation}
        P_{\mathcal{G}_v} (x|\theta_v) = \frac{1}{k} \sum_{i-1}^k \phi(x,\mathcal{G}_v(z_i))
   \approx \frac{1}{k} \sum_{i-1}^k exp(-L(x,\mathcal{G}_v(z_i))) ; z_i \sim P_z\mathrm{.}
\label{formula:ganleak}
\end{equation}
Based on this equation, the probability of the query instance $x$ being a member of the training dataset of the generative model is calculated based on its average distance to the generated synthetic data.

\textbf{\emph{Monte-Carlo attack}}~\cite{hilprecht2019monte}.
Similar to the DCR attack, the Monte-Carlo attack uses the distance to synthetic data points as a proxy for membership.
The intuition behind this attack is that if a generator is trained to generate instances close to training data, it will overfit. 
In this respect, the probability of a target point $x$ being a member increases if it is close to a generated point. 
To estimate this probability, an $\epsilon$-neighbourhood of point $x$ is defined as $U_{\epsilon}(x) = \{x'|d(x,x')\leq \epsilon\}$. 
The adversary looks into a ball with radius $\epsilon$ around the target point $x$ and counts the number of existing generated instances in this ball. 
According to the Monte-Carlo theory~\cite{owen2013monte}, the probability of a target point $x$ being a member of the training data of the generative model is computed as follows:
\begin{equation}
    \hat{f}_{MC-\epsilon}(x)=\frac{1}{n}\sum_{i=1} ^n 1_{g_i \in U_\epsilon(x)},
    \label{formula:monte-carlo}
\end{equation}
in which $g_i \in U_\epsilon(x)$ includes instances in the ball $U$ with radius $\epsilon$ around query instance $x$.
If this probability is higher than a predefined threshold, the target point is considered a member, while otherwise it is not the case. 

The \textbf{\emph{DOMIAS attack}}~\cite{van2023membership} also benefits from the generative model's overfitting to the training data in performing MIA.
An additional assumption is that the adversary has access to some auxiliary data sampled from the same distribution as the training data. 
Considering access to this data distribution, the attack formulation changes to the following equation, in which $p_R(X)$ refers to the real data distribution.
\begin{equation}
    A_{DOMIAS}(x^*) = f\left(\frac{p_G(x^*)}{p_R(x^*)}\right)\mathrm{.}
    \label{formula:domias}
\end{equation}
In Equation~\eqref{formula:domias}, $f$ is a monotonically increasing function with a range between zero and one, showing the probability of instance $ x^*$ being a member of the training data.
Thus, increasing the probability of the target instance belonging to the synthetic data distribution compared to that of belonging to the reference data distribution results in a higher membership score for the query instance.

The \textbf{\emph{Data Plagiarism Index (DPI)}}~\cite{ward2024data} evaluates the density ratio of synthetic data compared to the reference (auxiliary) data to analyze the local memorization around the target point. 
More precisely, DPI generates a $k$-neighbourhood($x*$) for each target point $x*$ using the synthetic and reference datasets. 
The DPI value $\rho$ is calculated as:
\begin{equation}
    \mathrm{\rho}(x^*) =
\frac{\sum_{z \in D(x^*)} \mathbb{I}(z \in S)}
     {\sum_{z \in D(x^*)} \mathbb{I}(z \in R)}\mathrm{.}
\end{equation}
When there is no synthetic data in the neighborhood ($DPI = 0$), it is a sign of underfitting.
In contrast, if the number of synthetic and reference data is equal ($DPI = 1$), no data plagiarism has happened, while a higher number of instances in synthetic data compared to the reference data ($DPI > 1$) is a sign of overfitting in the generative model, which leaks membership. 

The \textbf{\emph{Gen-LRA attack}}~\cite{ward2025privacy} builds a surrogate density estimator over $R$ (\emph{i.e.}, the reference set), which is used to estimate the likelihood of $S$ (\emph{i.e.}, the synthetic data). 
If the likelihood of $S$ is significantly higher in a density estimator over $R \cup x^*$, it illustrates overfitting.
An improvement on this attack is to localize this evaluation to samples close to $x^*$.
The membership score for the GenLRA is computed as:
\begin{equation}
    \mathrm{f_{Gen\text{-}LRA}}(x^*) =
\frac{\prod_{s \in S} p_{R \cup \{x^*\}}(s)}
     {\prod_{s \in S} p_{R}(s)}\mathrm{.}
\end{equation}

\textbf{\emph{LOGAN/Classifier}}~\cite{hayes2017logan}, trains a GAN using synthetic data to approximate the target's characteristics. 
The discriminator of this GAN learns to distinguish between reference data and synthetic data. 
This discriminator is then used to investigate the membership of the query instance in the training model, based on the assumption that the member instances will be classified as synthetic data~\cite{houssiau2022tapas}.

\section{Counterfactual generation mechanisms}
\label{sec:cfgeneration}

Counterfactuals are explanations of why some profiles received undesired decisions from the model~\cite{wachter2017counterfactual}. 
They have properties somewhat similar to those of synthetic data, which include proximity, plausibility and diversity~\cite{karimi2022survey}.
The \emph{proximity} means counterfactuals should be as close as possible to the query instance, \emph{plausibility} concerns generating instances that lie near the real data manifold, while finally \emph{diversity} addresses the concern about generating mutually distinct counterfactuals for each instance, showing various ways for changing the outcome.
Counterfactuals can be formally defined as follows: 
\begin{definition}
Given an input profile with feature values $x^o_1,\ldots,x^o_n$ and the corresponding model's prediction $d_1$, a counterfactual explanation method generates a counterfactual with feature values $cf_1,\ldots,cf_n$ satisfying two conditions: (1) the model should assign it a different prediction than from the original instance and (2) it should be close to the original instance in terms of a predefined distance, with the Euclidean distance being one of the most commonly used in counterfactuals.    
\end{definition}

Various techniques have been suggested to generate counterfactuals, which can be divided into two main categories: perturbation-based and instance-based counterfactuals.
Perturbation-based methods take the original instance and then perturb its feature values toward the decision boundary until the model's decision changes. 
The associated instance is then considered as the generated counterfactual.
While these methods generate counterfactuals with the lowest change in the original instance, they often suffer from a low level of plausibility~\cite{laugel2019dangers}. 
Instance-based counterfactuals are suggested to address this problem by using original instances from the training dataset to generate more realistic counterfactuals.
Hereafter, one method from each category is explained that will be later used in Section~\ref{chap:experimental-results} to assess the effectiveness of synthetic data MIA on counterfactuals generated using both techniques.

The \textbf{\emph{Nearest Instance Counterfactual Explanations (NICE)}}~\cite{brughmans2023nice} first identifies the nearest neighbour of the original instance for which the model makes a different prediction. 
Then, through an iterative process, the feature values of the factual instance are replaced with the values of the nearest neighbour until the model changes its prediction.
The selection of the feature values is based on a reward function, which integrates several criteria related to the quality of counterfactuals, namely proximity, sparsity and plausibility.

\textbf{\emph{Diverse Counterfactual Explanations (Dice)}} ~\cite{mothilal2020explaining} generates diverse actionable counterfactuals by solving an optimization algorithm for any differentiable model. 
More precisely, in addition to minimizing the distance between the original instance and its generated counterfactual, the objective of this optimization is to generate a number of diverse counterfactuals to give users the chance to decide how they want to update their profile to change the model's decision. 
The following equation formalizes this optimization problem, which is optimized through gradient descent:
\begin{equation}
C(x) = \arg\min_{c_1,\dots,c_k} \; 
\frac{1}{k}\sum_{i=1}^k yloss\!\bigl(f(c_i),y\bigr)
 + \frac{\lambda_1}{k}\, dist(c_i,x)
- \lambda_2\, dpp\_diversity(c_1,\dots,c_k) \mathrm{,}
\label{formula:dice}
\end{equation}
in which $c_i$ is a counterfactual explanation, $k$ is the number of counterfactuals, $yloss$ is a metric to minimize prediction error and $dpp\_diversity$ is the diversity parameter. 
$\lambda_1$ and  $\lambda_2$ are hyperparameters to balance the effects of proximity and diversity.  

\textbf{\emph {SCFE}} as suggested by Wachter and collaborators~\cite{wachter2017counterfactual}, is a gradient-based counterfactual generation algorithm like DICE. 
This method differs from other counterfactual mechanisms in that it treats all features as numerical features.
The distance mechanism used in this method is the following:
\begin{equation}
d(x, x')
=
\sum_{k \in F}
\frac{\left| x_k - x_k' \right|}{\mathrm{MAD}_k} \mathrm{,}
\label{eq:scfe-dist}
\end{equation}
in which
\begin{equation}
\mathrm{MAD}_k
=
\operatorname{median}_{j \in D}
\left(
\left| X_{j,k}
-
\operatorname{median}_{l \in P}(X_{l,k})
\right|
\right) \mathrm{.}
\end{equation}

\textbf{\emph{Instance-based Diverse Counterfactual Explanations (Dice-kdtree)}}. 
In addition to instance-based and perturbation-based counterfactual mechanisms, it is also possible to select an instance from the training dataset that is already classified in the counterfactual class and return it as the counterfactual. 
Mothiel and collaborators have proposed to fit a decision tree on the training dataset~\cite{mothilal2020explaining}. 
Then, for each query instance, the approach identifies the closest instance to the query by following the decision tree path and outputs that instance as the counterfactual.


\paragraph{Counterfactual distance attack (dist-lrt).}
Counterfactuals are generated as closely as possible to the original instances~\cite{wachter2017counterfactual} and, consequently, to the decision boundary.
Using these assumptions, the counterfactual distance attack~\cite{pawelczyk2023privacy} uses the distance between query instances and their counterfactuals as a proxy of the query instances' distance to the decision boundary.
Like in some other MIAs~\cite{shokri2017membership}, shadow models are used to simulate the target model properties.
More precisely, shadow models are surrogate models trained to imitate the target model's behaviour while providing the adversary with white-box access to their inner workings, training data and prediction vectors.

The counterfactual distance attack trains $n$ shadow models on shadow datasets that do not include the query instance.
Then, all shadow models are used to generate counterfactuals for the query instance.
Since the query instance has not been used in any of the shadow models' training sets, these distances are used to estimate the distribution of non-member counterfactual distances. 
Given this distribution and the query instance's counterfactual distance for the target model, the adversary predicts whether each query instance belongs to the model's training dataset. 
The full likelihood ratio is defined as:
\begin{equation}
\Lambda =
\frac{\Pr\!\left[c(x, x') \mid x \in D_t\right]}
     {\Pr\!\left[c(x, x') \mid x \notin D_t\right]},
\end{equation}
\noindent in which $c(x, x')$ is the distance between the counterfactual and the query instance, and $D_t$ is the training dataset.
Using diverse techniques to generate counterfactuals facilitates investigating the efficiency of our proposed attack compared to the baseline in relation to each specific technique. 


\section{Methodology}
\label{chap:methodology}
This section formalizes our threat models and details the experimental pipeline used to evaluate membership inference attacks (MIAs) on counterfactual explanations. 
We consider two attack categories: (i) a no-box MIA that treats released counterfactuals as a synthetic dataset and (ii) a counterfactual-distance baseline that requires query access to the target model.

\subsection{Threat models}
We define the two attacks with the following two-player security game between a challenger (model provider) and an adversary (attacker)~\cite{huang2011adversarial}.

\subsubsection{No-box MIA from released counterfactuals}
        \begin{enumerate}
            \item The challenger picks a sample dataset from the target distribution and generates a training dataset $\mathcal{D} \leftarrow \mathrm{D}$ and captures the target $z$ from the universe $\mathcal{U}$.
            \item The challenger trains the model $f_{\theta} \leftarrow \mathcal{T}(\mathcal{D})$.
            \item The challenger provides users with query access to the model $f_{\theta}$ and gives a counterfactual explanation $cf(x)$ for the undesired outcomes they receive for their query $x$.
            \item The adversary collects a set of counterfactuals $cf(x)$ generated for various users.
            \item The adversary has no access to the model $f_{\theta}$ or training set $\mathcal{D}$. 
            \item The adversary makes a guess $g$ based on their knowledge about the distribution $\mathrm{D}$ and the counterfactuals they have collected.
            \item The adversary wins if $g \in \mathcal{D}$.
        \end{enumerate}

\subsubsection{Counterfactual-distance baseline (query access)}
\begin{enumerate}
            \item The challenger picks a sample dataset from the target distribution and generates a training dataset $\mathcal{D} \leftarrow \mathrm{D}$ and captures the target $z$ from the universe $\mathcal{U}$.
            \item The challenger trains the model $f_{\theta} \leftarrow \mathcal{T}(\mathcal{D})$.
            \item The challenger provides users with query access to the model $f_{\theta}$ and gives a counterfactual explanation $cf(x)$ for the undesired outcomes they receive for their query $x$.
            \item The adversary gets $x' = cf(x)$ from model $f_{\theta}$ and calculates distance between $x$ and $x'$ as $c(x,x')$.
            \item The adversary selects a false positive rate $\alpha \in (0,1)$.
            \item The adversary samples $N$ set of instances $\mathcal{D_t^i}$ from dataset $\mathcal{D}$ and trains $N$ shadow models on them. $\mathcal{D_t^i}$ is considered not having instance $x$.
            \item The adversary generates counterfactuals for query instance $x$ using all $N$ shadow models and gets counterfactual distances $\{c(x, x'^{(1)}), \ldots, c(x, x'^{(N)})\}$.
            \item The adversary calculates the maximum likelihood estimates for the mean ($\hat{\mu}_{\mathrm{MLE}}$ and variance $\hat{\sigma}^2_{\mathrm{MLE}}$ of these shadow distances.
            \item The adversary compares the target distance $c(x,x')$
            against this estimated distribution.
            if $x(x,x')$ is bigger than $Z_{1-\alpha}$ quintile of the non-member distribution, the guess $g$ is non-members. 
            Otherwise, $g$ is a member.
            \item The adversary wins if $g$ is correct.
        \end{enumerate}
        
\subsection{Attack framework} \label{sec:attackframework}
\begin{figure*}[h!]
\centering
\includegraphics[width=.8\linewidth]{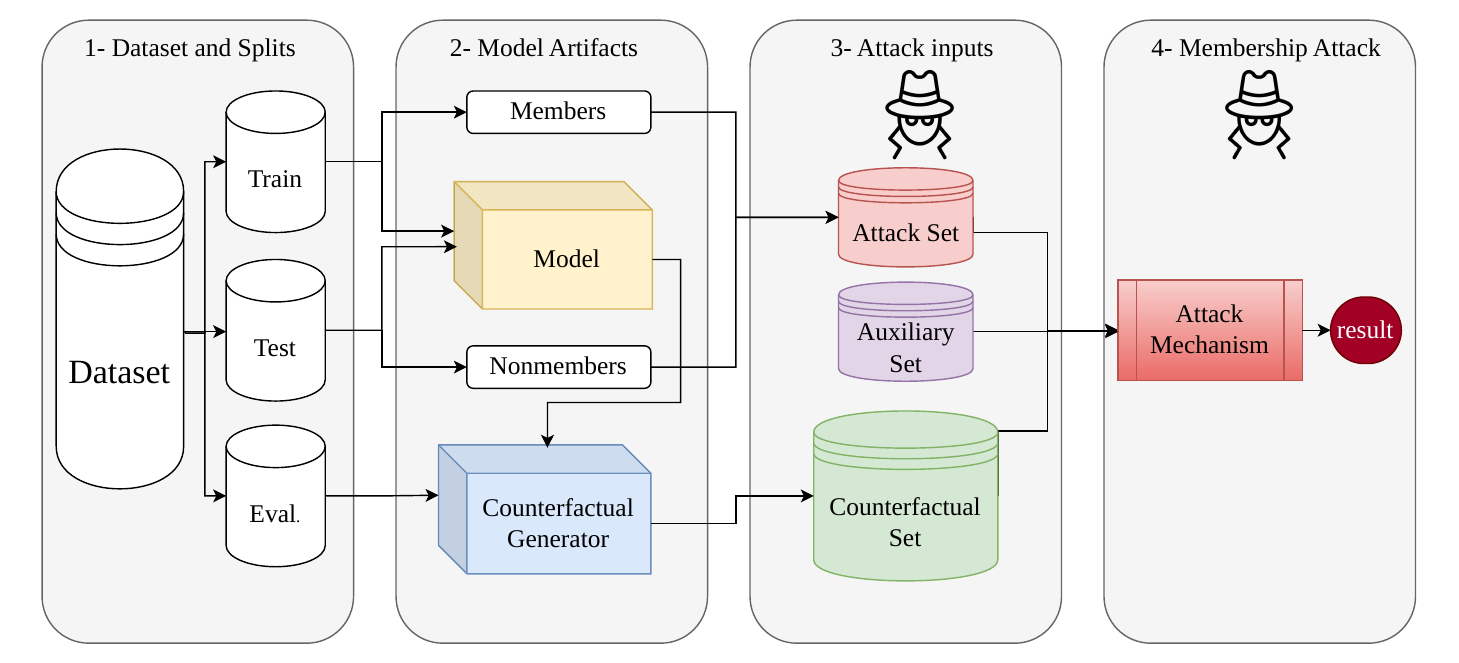}
\caption{Attack pipeline. The attacker has no access to or information about the model or counterfactual generator. They only have a set of counterfactuals, a small auxiliary set from the same distribution of the dataset, and a set of query instances to find their membership status.}
\Description{attack pipeline}
\label{fig:attack-pipeline}
\end{figure*}

\autoref{fig:attack-pipeline} presents a high-level overview of the attack pipeline.
Our evaluation follows a common pipeline across datasets and counterfactual generation mechanisms. 
For each dataset, we split the data into training (60\%), test (20\%) and counterfactual (20\%) sets. 
We first train the target model on the training set before evaluating it on the test set. 
Finally, counterfactuals were generated for instances drawn from the counterfactual split.
These counterfactuals constitute the ``synthetic'' dataset available to the no-box adversary.
The default synthetic set size for the attack is 10000 instances. 
Details on the implementations and sizes are provided in Section~\ref{chap:experimental-results}.

Following the approach used by Ward and collaborators~\cite{ward2025ensembling} to build an attack dataset, we sample an equal number ($n=500$) of members from the training set and non-members from the test set (ensuring the non-member set contains no training instances via re-identification checks).
Additionally, we sample a reference set from the test set following the same general protocol as non-members, which is utilized by some of the individual attacks in the ensemble, compared to the counterfactual distances baseline attack, which requires inference access to the model's predictions and counterfactuals.

Following our pipeline in Figure~\ref {fig:attack-pipeline}, synthetic and attack datasets are fed into the attack model. 
More precisely, for the no-box setting, we apply an ensemble of MIAs originally designed for synthetic tabular data using the collected counterfactuals as the synthetic dataset. 
The implemented attacks include Distance to Closest Record DCR/DCR-Diff, DOMIAS, Data Plagiarism Index (DPI), Gen-LRA, LOGAN/Classifier, and Monte Carlo (MC). 
Each attack outputs a membership score per target instance and we then apply ensembling to combine signals across attacks.
In particular, the following ensembling strategies are used:
\begin{itemize}
    \item \emph{Majority voting}. Each instance is considered a member by majority voting if the majority of individual attacks decide it is a member with confidence larger than a predefined threshold. 
    We followed~\cite{ward2025ensembling}, which used majority voting with thresholds ranging from 90\% to 98\%.
    \item \emph{Mean aggregation}. The membership score assigned to the instance is the mean score assigned to it by all individual attacks.
\end{itemize}

While ensembling does not always outperform all individual attacks, it typically achieves better performance than using any single individual attack.
According to Ward and collaborators~\cite{ward2025ensembling}, ensembling works well because each individual attack performs better on some synthetic data depending on the dataset properties and the algorithm used to generate it. 
Therefore, ensembling helps strong attacks to influence the final prediction more than weak attacks, which is helpful because the adversary does not know how synthetic data (in our case, counterfactuals) are generated or which attack will perform better on the data.

\section{Experimental evaluation} 
\label{chap:experimental-results}
The experimental results of the attacks are provided in this section. 
To provide reproducibility, our code is available at \href{https://github.com/tisl-lab/Counterfactual-distance-attack}{Counterfactual distance attack} and \href{https://github.com/tisl-lab/Nobox-MIA-CF.git}{No-box ensemble MIA}. Both repositories have an execution guide.ipynb file. 
All results presented here, including counterfactual information, the model's accuracy and attack results for no-box and baseline attack, are averaged over five individual executions using different random seeds. \

\subsection{Experimental setup} \label{sec:experimental-setup}

\paragraph{Datasets}
We performed our attack on four tabular datasets (summarized in \autoref{tab:datasets}), including Adult income, Acs\_income, Compas and Heloc.
\textbf{Adult income}~\citep{asuncion2007uci} includes information of 48842 individuals that are described with 14 features, with the learning task being to predict whether the profile owner earns more than 50k.
Following the literature, we dropped five features (education-num, fnlwgt, native-country, capital-gain and capital-loss) that are redundant or not related to the training task, and all the experiments used 9 related features. 
\textbf{Acs\_income} is the new version of the Adult dataset, with 10 features and a training set of 1.66M instances. 
In our implementation, after preprocessing and removing instances with missing feature values, the final dataset size is 199,665 instances.
\textbf{Compas}~\cite{angwin2016machine} is a dataset used for recidivism prediction composed of 6172 profiles described by 8 features. 
The prediction task of this dataset is to predict the risk of recidivism. 
\textbf{Heloc}~\cite{openml_heloc_45554}, or home equity line of credit, uses 23 features to predict whether a profile is high-risk or low-risk of approval for the line of credit.
An overview of the datasets and the accuracy of models trained on them is presented in Table~\ref{tab:datasets}. 

\begin{table}[ht]
\centering
{
\begin{tabular}{lllll}
\hline
 & \textbf{Adult} & \textbf{Compas} & \textbf{Heloc} & Acs\_income\\
\hline

\rowcolor{lightgray}
\#features 
& 9 & 8 & 23 & 10\\

\hline
Dataset size 
& 48842 & 6172 & 10000 & 199665 \\

\hline
\rowcolor{lightgray}
Class 
& Income & Low\_risk & Riskperformance & Income\\

\hline
Accuracy
& 81\% & 68\% & 72\% & 80\% \\

\hline
\end{tabular}
}
\caption{Summary of datasets characteristics and accuracy of their models.}
\label{tab:datasets}
\end{table}





 %
\label{exp:ablation-studies}
To determine the appropriate size of the attack and synthetic set sizes, we performed two sets of ablation studies. 
The corresponding results are provided in~\autoref{app:ablation-atudies}.
Our findings indicate that for small datasets such as Compas and Heloc, relatively small attack set sizes of 200 to 400 instances are sufficient to achieve optimal attack performance. 
For these datasets, a synthetic (counterfactual) set size of 500 to 1000 is required to optimize the attack performance.
For larger datasets, both a larger attack set and a higher number of counterfactuals are necessary to optimize the attack. 
In particular, an attack dataset of 1000 instances is needed for adult and $acs\_income$ to obtain the most effective attack. 
For these datasets, the counterfactual set must contain 5000 to 10000 instances to reach the highest performance.
\autoref{tab:ablation_sample} shows a sample of this ablation study for $Acs\_income$, showing how increasing attack set size improves attack performance. 
More detailed results are presented in~\autoref{app:ablation-atudies}.

\begin{table*}[ht]
\centering
\begin{tabular}{lllll}
\hline
 \textbf{Synth Size} & \textbf{ROC AUC} & \textbf{$\mathbf{TPR@FPR.01}$} & \textbf{$\mathbf{TPR@FPR.1}$} & \textbf{PR AUC} \\
\hline
\rowcolor{lightgray}
500 & 0.505 & 0.011 & 0.103 & 0.505 \\
 1000 & 0.503 & 0.016 & 0.110 & 0.508 \\
\rowcolor{lightgray}
2000 & 0.502 & 0.016 & 0.113 & 0.510 \\
5000 & 0.505 & 0.017 & 0.120 & 0.512 \\
\rowcolor{lightgray}
10000 & 0.506 & 0.016 & 0.117 & 0.512 \\
\hline

\hline
\end{tabular}
\caption{$Acs\_Income$ - Fixed Attack Set Size 1000.}
\label{tab:ablation_sample}
\end{table*}

Having the results of the ablation study, to perform a MIA, we generated up to 10000 counterfactuals using each method for instances in the counterfactual set explained in Section~\ref{fig:attack-pipeline}.
For smaller datasets, such as Heloc and Compas, which have fewer instances, both the synthetic and attack datasets are smaller 
since there are fewer instances in the counterfactual set to generate counterfactuals, and fewer instances in the test set to be used as non-members.
For Compas, the synthetic dataset size is $1443$ instances and the attack set size contains $500$ instances.
Consequently, for Heloc, the attack set includes $1995$ instances, while the attack set size maintains the size of $1000$.
To keep consistency and fair comparison, for the baseline attack, \emph{i.e.}, counterfactual distance attack, the same attack set size of $1000$ instances has been used to generate counterfactuals and perform the attack. 
While the no-box ensemble attack has access to a reference set of 500 instances sampled from non-member instances of the training distribution, the baseline counterfactual distance attack has additional access to the model's predictions and counterfactuals.

\paragraph{Counterfactual methods} \label{exp:methodselection}
We implemented four counterfactual mechanisms to generate counterfactuals.
Since our no-box attack is not limited to any synthetic data generation mechanism, we did not limit counterfactual generation mechanisms as well. 
To be more precise, since various attacks are ensembled in our setting, each individual attack can take advantage of some of the counterfactual generation mechanisms and their properties and improve the ensemble results. 
Thus, we implemented various techniques to see how this attack works on counterfactuals generated using each of them, without limiting the counterfactual methods used. 

These generated counterfactuals are used in our no-box attack setting, in which the adversary, lacking access to the original model, uses them as synthetic data to perform a membership inference attack.  
The counterfactual mechanisms implemented are Nice, a representative of instance-based counterfactuals, Dice\_gradient and SCFE, representatives of perturbation-based counterfactuals, as well as Dice-kdtree, used as a sanity check since all counterfactuals produced are members of the training dataset. 
The reason why we have implemented two perturbation-based counterfactuals is their use of different encoding for feature values, which affects the final counterfactuals generated.
dice\_gradient uses one-hot encoding for categorical features and standard scaler for numerical features, with the tendency to perturb numerical features, while SCFE treats all features as numerical values using standard scalers for them. 
\label{exp:cf-setting}
Among counterfactual methods we used, $dice\_gradient$ has the potential for setting various hyperparameters to prioritize different objectives, including proximity, diversity, sparsity and actionability. 
The parameters used for the results presented in the main paper are as follows:
$proximity\_weight: 0.1$, $diversity\_weight:1.0$ and all features are allowed to change during the counterfactual generation mechanism. 
This setting has been used to achieve fair results compared to the baseline attack, without tailoring counterfactuals toward a more vulnerable setting to our no-box attack.
The statistical analysis of counterfactuals generated using each method is presented in~\autoref {tab:cf-stats}.



\begin{table*}[ht]
\centering
{
\begin{tabular}{lllrr}
\toprule
\textbf{Dataset} & \textbf{CF} & \textbf{avg\_distance ± std} & \textbf{avg\_reid\_rate ± std} & \textbf{success\_rate} \\
\midrule

\rowcolor{lightgray}
acs\_income & NICE & 0.355 ± 0.311 & 0.108 ± 0.320 & 100.000 \\
 & dice\_gradient & 0.783 ± 0.398 & 0.000 ± 0.000 & 99.967 \\
\rowcolor{lightgray}
 & dice\_kdtree & 1.104 ± 0.476 & 1.011 ± 0.125 & 100.000 \\
 & scfe & 0.538 ± 0.623 & 0.000 ± 0.000 & 99.999 \\

\hline

\rowcolor{lightgray}
adult & NICE & 0.922 ± 0.848 & 0.760 ± 1.302 & 100.000 \\
 & dice\_gradient & 1.346 ± 0.415 & 0.000 ± 0.000 & 99.947 \\
\rowcolor{lightgray}
 & dice\_kdtree & 1.085 ± 0.573 & 1.222 ± 0.666 & 100.000 \\
 & scfe & 0.434 ± 0.632 & 0.000 ± 0.000 & 99.994 \\

\hline

\rowcolor{lightgray}
compas & NICE & 0.194 ± 0.268 & 3.019 ± 4.849 & 100.000 \\
 & dice\_gradient & 0.824 ± 0.315 & 0.001 ± 0.042 & 99.945 \\
\rowcolor{lightgray}
 & dice\_kdtree & 0.388 ± 0.456 & 3.451 ± 3.710 & 99.945 \\
 & scfe & 0.014 ± 0.071 & 0.000 ± 0.000 & 100.000 \\

\hline

\rowcolor{lightgray}
heloc & NICE & 0.298 ± 0.305 & 0.000 ± 0.009 & 100.000 \\
 & dice\_gradient & 1.132 ± 0.315 & 0.000 ± 0.000 & 100.000 \\
\rowcolor{lightgray}
 & dice\_kdtree & 0.841 ± 0.471 & 0.810 ± 0.391 & 100.000 \\
 & scfe & 0.073 ± 0.168 & 0.000 ± 0.000 & 91.635 \\

\bottomrule
\end{tabular}
}
\caption{Counterfactuals statistic analysis across datasets and methods. $avg\_distance$ is the average distance of generated counterfactuals to the query instance, $avg\_reid\_rate$ shows the average number of exact matches existing in the training dataset for each counterfactual. $success\_rate$ shows the percentage of queries for which the method successfully generated a valid counterfactual.}
\label{tab:cf-stats}
\end{table*}

\paragraph{Baseline attack}
To evaluate the effectiveness of the ensemble-MIA attack designed for synthetic data, we implemented the counterfactual distance attack~\cite{pawelczyk2023privacy}, which relies on counterfactual distances and their distributions for members and non-members to infer whether a target instance belongs to the training dataset.
This attack requires query access to the model, as the adversary must be able to request counterfactuals for query instances to compute counterfactual distance and compare it with the distributions of distances generated by the shadow models.
Due to different operational requirements, we run this attack in a separate pipeline, using the same datasets and the same counterfactual metrics.
To the best of our knowledge, it remains the only MIA attack implemented on counterfactuals generated for tabular datasets.

\paragraph{Evaluation metrics}
To evaluate the performance of the attacks, we have used the following metrics in accordance with the literature~\cite{shokri2017membership,hilprecht2019monte,chen2020gan,hu2021tablegan,van2023membership,ward2025ensembling}:
\begin{itemize}
    \item Access level (white-box, black-box, no-box).
    \item Shadow models training required.
    \item TPR for fixed small FPR: True positive rate at low false positive rate, measuring the ability to correctly identify members while maintaining a low false alarm rate. 
    \item ROC AUC: The overall attack ability to distinguish members and non-members across all attack datasets, independent of any decision threshold.
    \item Precision-Recall (PR) curve: Evaluates the trade-offs between successful member identification (proportion of samples identified as members that are actually members) and overall attack coverage (proportion of true members correctly identified).
\end{itemize}

\begin{table*}[ht]
\centering
\begin{tabular}{lllllll}
\hline
\textbf{Dataset} & \textbf{CF} & \textbf{Attack} & \textbf{ROC AUC} & $\mathbf{TPR@FPR.01}$ & \textbf{$\mathbf{TPR@FPR.1}$} & \textbf{PR AUC} \\
\hline

\rowcolor{lightgray}
Compas & Nice & Ensemble & \textbf{0.615} & \textbf{0.011} & \textbf{0.175} & \textbf{0.599} \\
 &  & Dist Lrt & 0.496 & 0.009 & 0.103 & 0.500 \\

\rowcolor{lightgray}
Compas & dice\_kdtree & Ensemble & \textbf{0.624} & \textbf{0.012} & \textbf{0.192} & \textbf{0.607} \\

 &  & Dist Lrt & 0.495 & 0.011 & 0.109 & 0.501 \\

\rowcolor{lightgray}
Compas & gradient & Ensemble & 0.602 & 0.019 & 0.200 & 0.592 \\
 &  & Dist Lrt & \textbf{0.652} & \textbf{0.023} & \textbf{0.328} & \textbf{0.643} \\

\rowcolor{lightgray}
Compas & scfe & Ensemble & \textbf{0.694} & \textbf{0.029} & \textbf{0.337} & \textbf{0.683} \\
 &  & Dist Lrt & 0.572 & 0.010 & 0.144 & 0.551 \\

\hline

\rowcolor{lightgray}
Heloc & Nice & Ensemble & \textbf{0.515} & \textbf{0.019} & \textbf{0.128} & \textbf{0.522} \\
 &  & Dist Lrt & 0.482 & 0.010 & 0.092 & 0.489 \\

\rowcolor{lightgray}
Heloc & dice\_kdtree & Ensemble & \textbf{0.534} & \textbf{0.047} & \textbf{0.164} & \textbf{0.555} \\
 &  & Dist Lrt & 0.481 & 0.015 & 0.111 & 0.501 \\

\rowcolor{lightgray}
Heloc & gradient & Ensemble & \textbf{0.491} & \textbf{0.009} & 0.093 & \textbf{0.496} \\
 &  & Dist Lrt & 0.471 & 0.006 & \textbf{0.098} & 0.477 \\

\rowcolor{lightgray}
Heloc & scfe & Ensemble & \textbf{0.495} & \textbf{0.009} & \textbf{0.095} & \textbf{0.498} \\
 &  & Dist Lrt & 0.472 & 0.004 & 0.093 & 0.476 \\

\hline

\rowcolor{lightgray}
Adult & Nice & Ensemble & \textbf{0.514} & \textbf{0.017} & \textbf{0.134} & \textbf{0.524} \\
 &  & Dist Lrt & 0.497 & 0.011 & 0.104 & 0.502 \\

\rowcolor{lightgray}
Adult & dice\_kdtree & Ensemble & \textbf{0.509} & \textbf{0.031} & \textbf{0.131} & \textbf{0.527} \\
 &  & Dist Lrt & 0.495 & 0.014 & 0.101 & 0.503 \\

\rowcolor{lightgray}
Adult & gradient & Ensemble & \textbf{0.512} & 0.011 & \textbf{0.116} & \textbf{0.511} \\
 &  & Dist Lrt & 0.490 & 0.011 & 0.097 & 0.492 \\

\rowcolor{lightgray}
Adult & scfe & Ensemble & \textbf{0.513} & 0.007 & \textbf{0.120} & \textbf{0.512} \\
 &  & Dist Lrt & 0.487 & \textbf{0.009} & 0.062 & 0.479 \\

\hline

\rowcolor{lightgray}
Acs\_income & Nice & Ensemble & \textbf{0.511} & \textbf{0.015} & \textbf{0.115} & \textbf{0.512} \\
 &  & Dist Lrt & 0.506 & 0.011 & 0.105 & 0.500 \\

\rowcolor{lightgray}
Acs\_income & dice\_kdtree & Ensemble & \textbf{0.510} & \textbf{0.017} & \textbf{0.121} & \textbf{0.519} \\
 &  & Dist Lrt & 0.497 & 0.014 & 0.094 & 0.499 \\

\rowcolor{lightgray}
Acs\_income & gradient & Ensemble & \textbf{0.505} & \textbf{0.018} & \textbf{0.112} & \textbf{0.511} \\
 &  & Dist Lrt & 0.504 & 0.013 & 0.108 & 0.510 \\

\rowcolor{lightgray}
Acs\_income & scfe & Ensemble & \textbf{0.513} & \textbf{0.013} & \textbf{0.107} & \textbf{0.515} \\
 &  & Dist Lrt & 0.496 & 0.002 & 0.061 & 0.476 \\

\bottomrule
\end{tabular}
\caption{Comparison of counterfactual-based attacks. Results are averaged over five different random seeds. It is worth mentioning that two differences between the Ensemble and Dist Lrt attacks are in the access level and shadow model training, where the Ensemble attack is no-box with no access to the model or training data, and the Dist Lrt has query access to the model and trains shadow models, which increases the complexity of the attack.}
\label{tab:cf_attacks}
\end{table*}
\subsection{Experimental results} \label{sec:experimental-results}
\begin{figure}[ht]
\centering
\begin{tabular}{c c c}

 & \textbf{No-box attack} & \textbf{CF-distance-attack} \\[4pt]

\rotatebox{90}{\textbf{Compas}}
&
\begin{minipage}{0.42\textwidth}
    \centering
    \includegraphics[width=\linewidth]{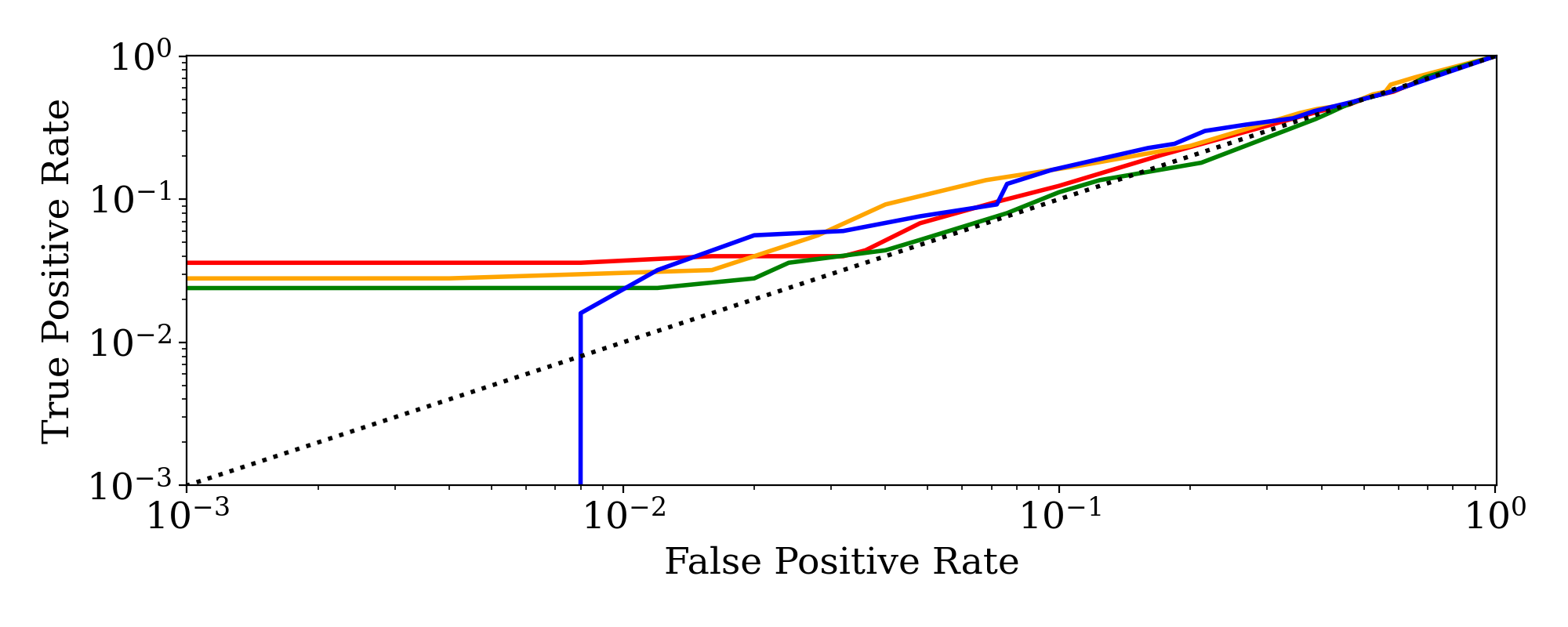}
\end{minipage}
&
\begin{minipage}{0.42\textwidth}
    \centering
    \includegraphics[width=\linewidth]{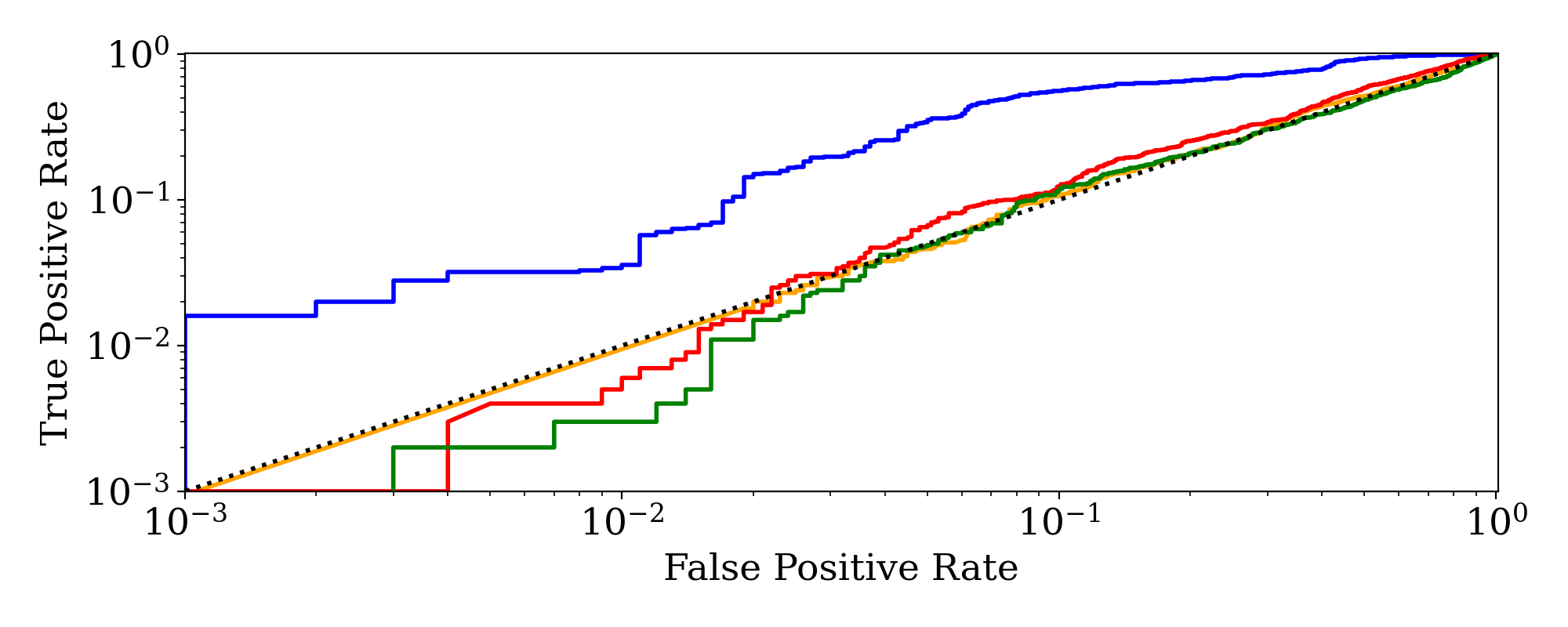}
\end{minipage}
\\[8pt]

\rotatebox{90}{\textbf{Heloc}}
&
\begin{minipage}{0.42\textwidth}
    \centering
    \includegraphics[width=\linewidth]{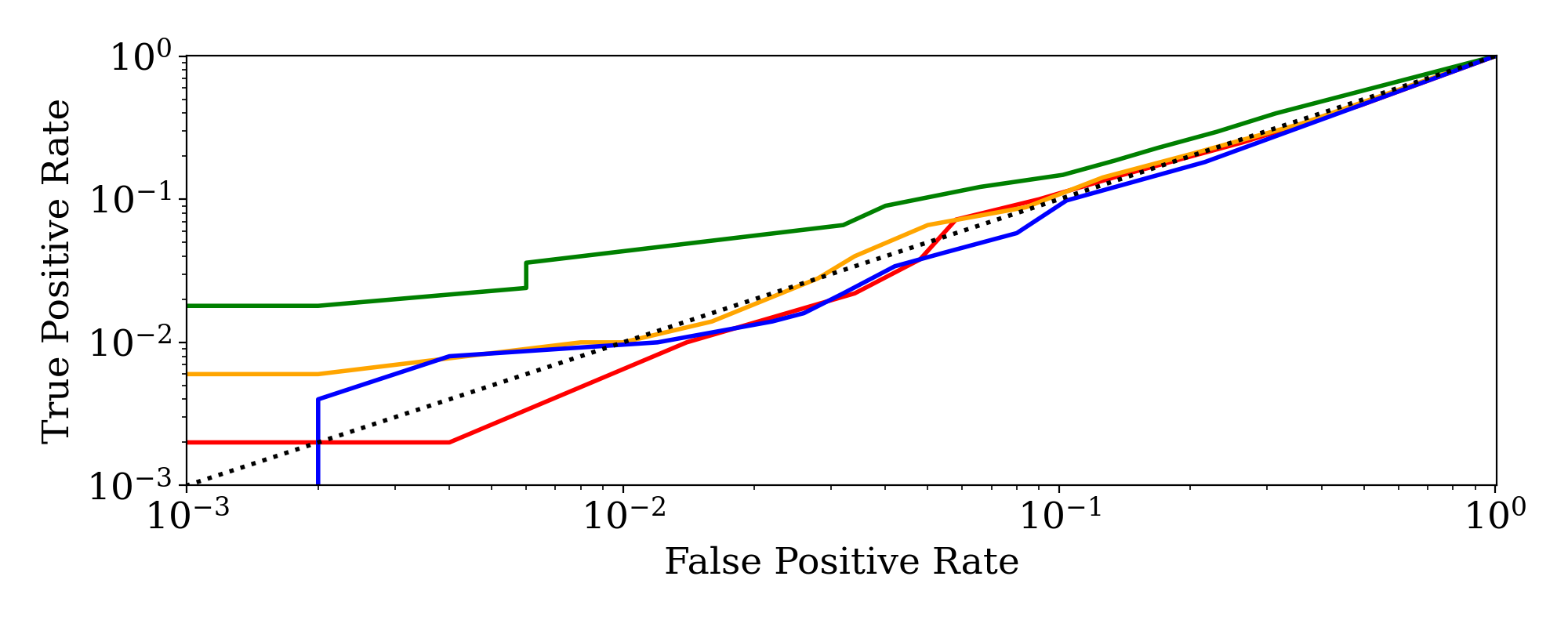}
\end{minipage}
&
\begin{minipage}{0.42\textwidth}
    \centering
    \includegraphics[width=\linewidth]{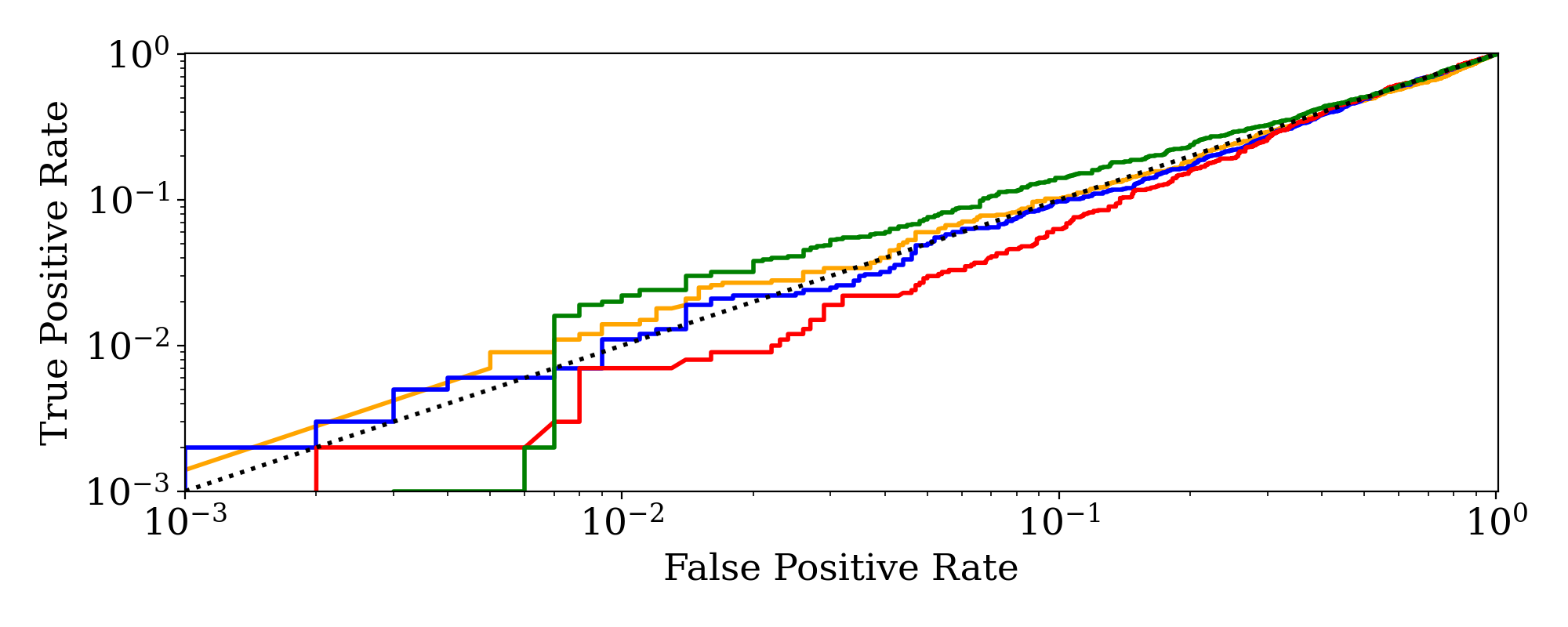}
\end{minipage}
\\[8pt]

\rotatebox{90}{\textbf{Adult}}
&
\begin{minipage}{0.42\textwidth}
    \centering
    \includegraphics[width=\linewidth]{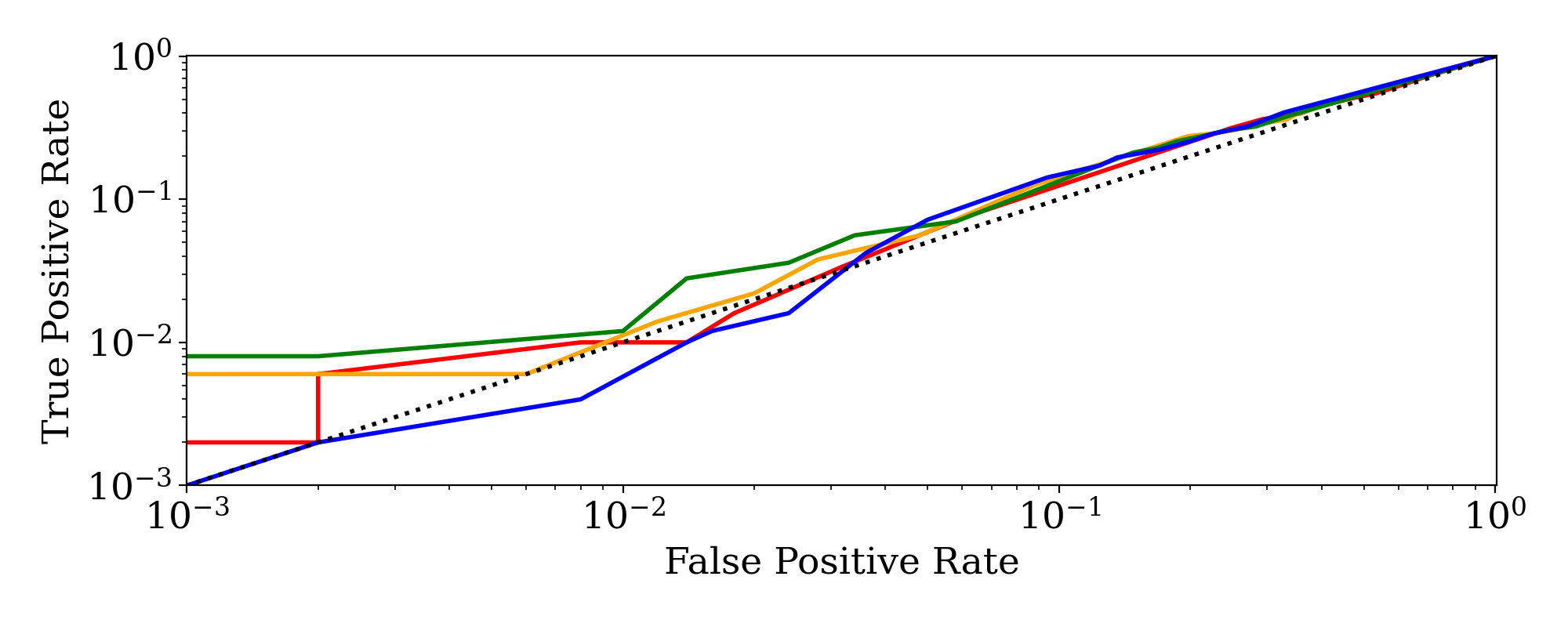}
\end{minipage}
&
\begin{minipage}{0.42\textwidth}
    \centering
    \includegraphics[width=\linewidth]{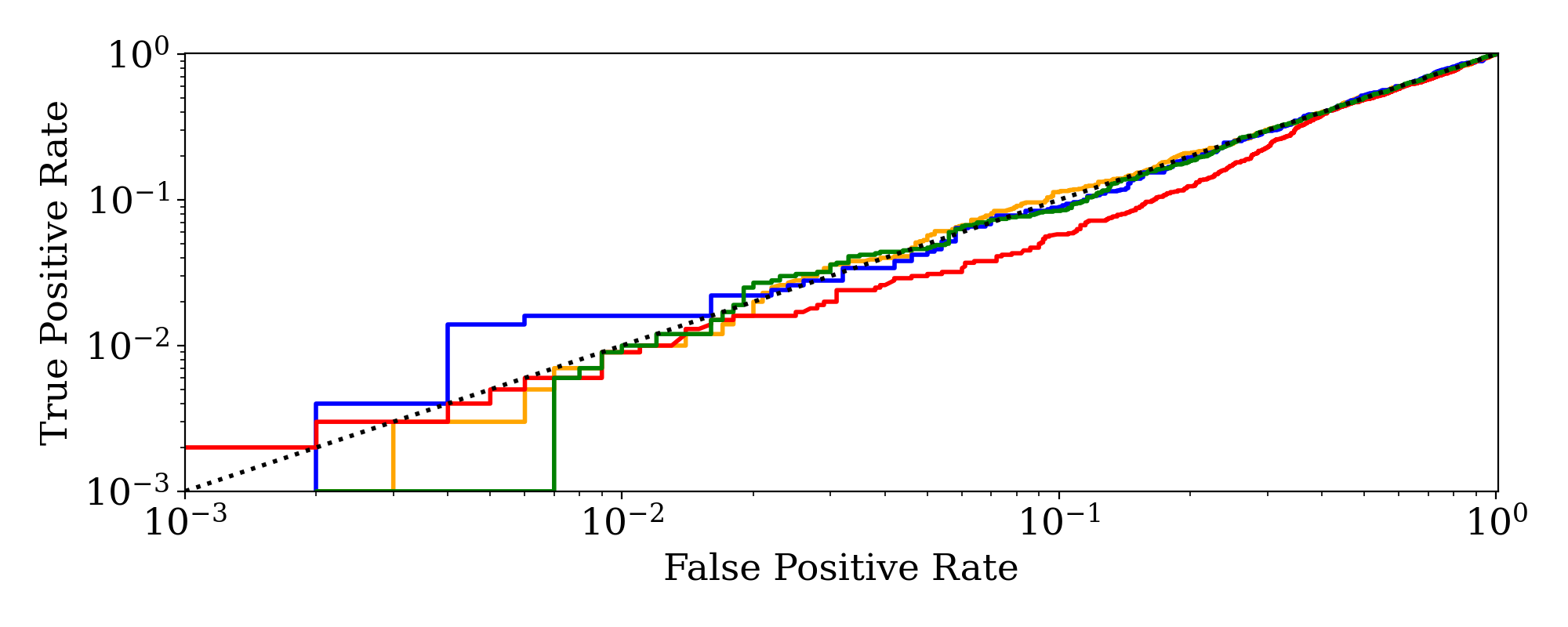}
\end{minipage}
\\[8pt]

\rotatebox{90}{\textbf{Acs Income}}
&
\begin{minipage}{0.42\textwidth}
    \centering
    \includegraphics[width=\linewidth]{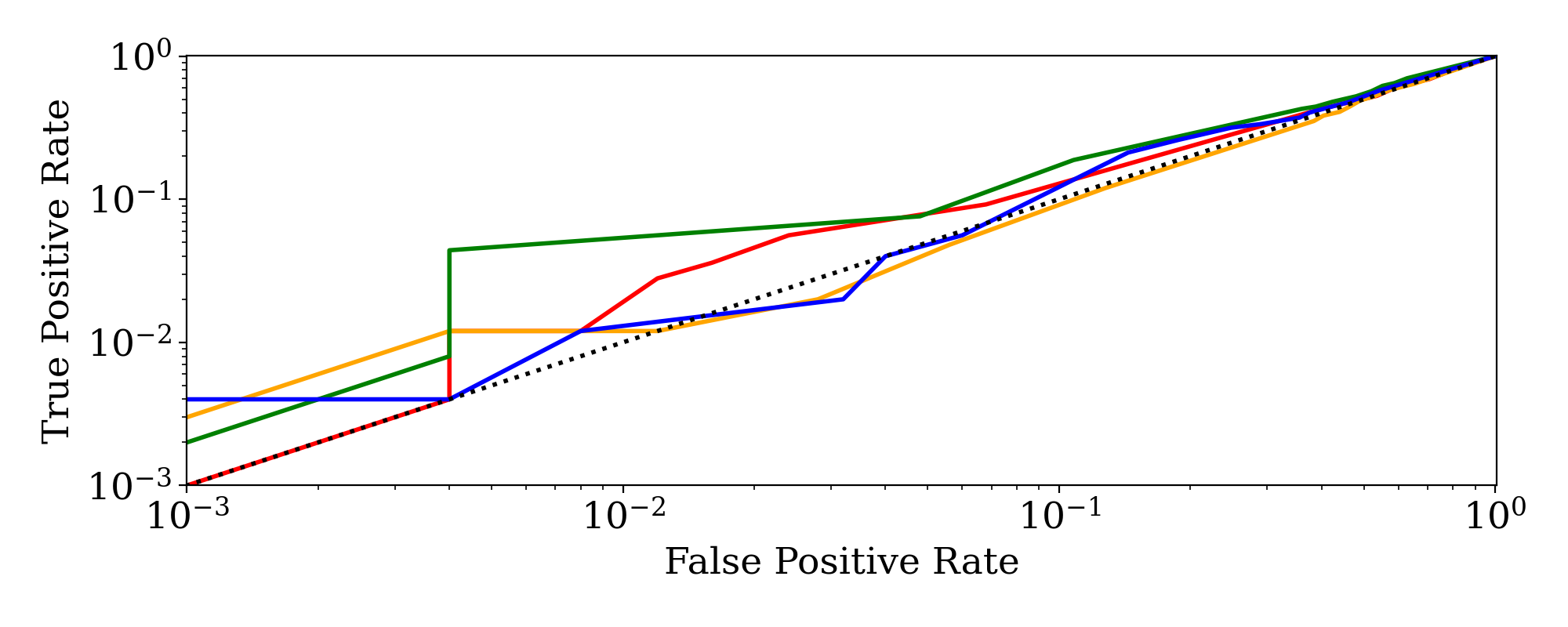}
\end{minipage}
&
\begin{minipage}{0.42\textwidth}
    \centering
    \includegraphics[width=\linewidth]{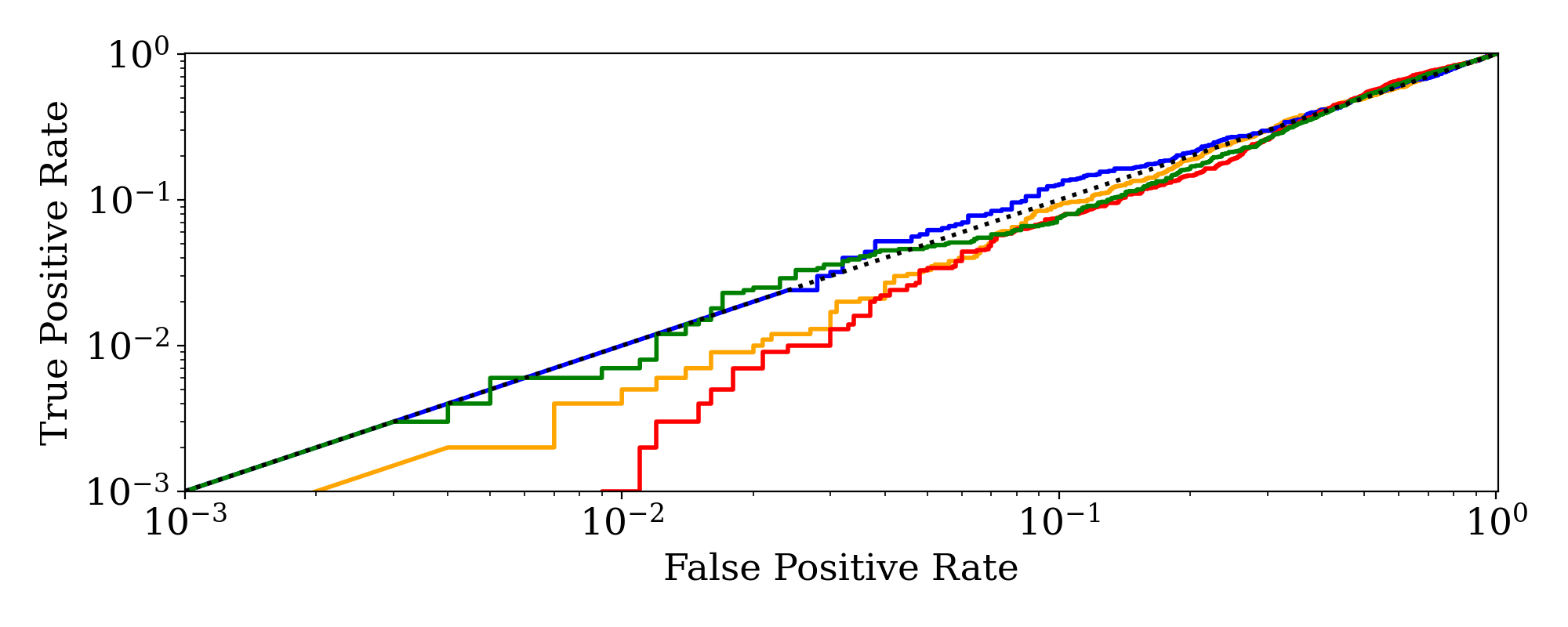}
\end{minipage}

\end{tabular}

\caption{Attack performance comparison. Blue lines show ROC curves for \textcolor{blue}{\texttt{dice\_gradient}}, green lines for \textcolor{darkgreen}{\texttt{dice\_kdtree}}, orange lines for \textcolor{orange}{\texttt{NICE}}, and red lines for \textcolor{red}{SCFE}. While No-box attacks work better on instance-based counterfactuals, the counterfactual distance attack is more accurate in the case of perturbation-based counterfactuals. Still, except for the Compas dataset, the counterfactual distance attack performs a bit better than a random guess.}
\label{fig:attack_performance}
\end{figure}

We evaluate the no-box ensemble MIA~\cite{ward2025ensembling} (originally designed for synthetic data) against the counterfactual distance attack (Dist-LRT) 
baseline~\cite{pawelczyk2023privacy}, with both attacks run on a server with 12 GB of RAM. \autoref{fig:attack_performance} and ~\autoref{tab:cf_attacks} summarize the performance of both attacks.

Overall, the ensemble MIA is consistently stronger than Dist-LRT in nearly all settings, with the main exception being dice\_gradient on Compas. 
These results suggest that meaningful membership leakage can arise even without any model-query access when counterfactuals are released. 
More specifically, the ensemble MIA outperforms the distance-based attack on instance-based counterfactuals (\emph{i.e.}, dice\_kdtree and NICE). 
In contrast, both attacks exhibit near-random-guess performance on perturbation-based methods (\emph{i.e.}, dice\_gradient and SCFE), with the ensemble MIA performing slightly better than the distance-based attack. 
This suggests that counterfactual realism or proximity to training points increases vulnerability to MIAs tailored for synthetic data. 

To support this claim, we performed a comparison of the distributions between the training and counterfactual sets for all CF generation mechanisms, presented in~\autoref{app:distribution-comparison}. 
This comparison shows that the membership inference attack performs better when counterfactual distributions are more similar to the training set. According to the distribution comparison, instance-based methods (i.e., $dice\_kdtree$ and \texttt{NICE}) generate CFs with higher similarity to the train set, which makes them more vulnerable to MIAs against synthetic data. In contrast, perturbation-based methods (i.e., $dice\_gradient$ and \texttt{SCFE}) change the distribution and thus reduce the performance of the ensemble attack.

We also conducted another set of experiments to evaluate the effect of proximity, diversity and actionability of the counterfactuals to their vulnerability to the membership inference attack. 
By setting various hyperparameters for $dice\_gradient$, we generated counterfactuals with focus on each of these hyperparameters to investigate how the attack performance changes. \autoref{tab:dicevariation_sample} provides a synthesized overview of these experiments, while comprehensive results are reported in \autoref{app:hyperparameters}. Overall, the results of these experiments show that:
\begin{itemize} \label{exp:hyperparameters}
    \item When no limitation on changeable features is applied, higher proximity slightly increases vulnerability to membership inference attack when proximity increases from $0.5$ to $1$, but increasing this hyperparameter more than $1$ reduces this vulnerability. 
    \item Limiting the changeable features to the actionable features only increases vulnerability to the membership inference attack for $acs\_income$, while decreasing the vulnerability for datasets such as Compas with more difficult classification tasks with higher dimensions.
\end{itemize}
\begin{table*}[ht]
\centering
\begin{tabular}{llllll}
\hline
\textbf{CF} & \textbf{ROC AUC} & \textbf{$\mathbf{TPR@FPR.01}$} & \textbf{$\mathbf{TPR@FPR.1}$} & \textbf{PR AUC} \\
\hline
dice\_gradient\_pw0.5\_dw0.1\_ftv-all & 0.502 & 0.010 & 0.106 & 0.506 \\
\hline
\rowcolor{lightgray}
dice\_gradient\_pw0.5\_dw0.1\_ftv\_actionable & 0.519 & \textbf{0.029} & 0.098 & 0.518 \\
\hline
dice\_gradient\_pw0.5\_dw0.5\_ftv-all & 0.502 & 0.010 & 0.106 & 0.506 \\
\hline
\rowcolor{lightgray}
dice\_gradient\_pw1\_dw0.1\_ftv\_actionable & 0.506 & 0.011 & 0.105 & 0.506 \\
\hline
\end{tabular}
\caption{$Acs\_Income$ - evaluation of attack performance across various proximity/diversity/actionability settings for Dice Gradient mechanisms.}
\label{tab:dicevariation_sample}
\end{table*}
Since the effect of various hyperparameter settings is not consistent among all datasets, we keep this direction open for more investigations.

Another set of experiments we performed is the evaluation of the attack performance based on the distance to the decision boundary.
More precisely, we divided query instances into five different bins based on their distance to the decision boundary.
We used confidence score as a proxy for this distance~\cite{shokri2021privacy,pawelczyk2023privacy}. The more confident the prediction is, the further the instance is from the decision boundary.
As expected, our results show that the membership inference attack is more successful for instances closer to the decision boundary.
The results of these experiments are provided in~\autoref{app:decision-boundary}
Finally, we observe that attack performance is higher on smaller datasets such as Compas and Heloc, which is consistent with prior evidence~\cite{tobaben2025impactdatasetpropertiesmembership} that smaller datasets can yield more memorization and thus higher MIA risk.

To explain why ensembling is more effective than individual attacks on counterfactuals in a no-box setting, we investigated the effectiveness of the individual attacks on different datasets and counterfactual generation mechanisms, and compared their effectiveness with the ensemble attacks. \label{exp:individuals}
The results (\emph{see}~\autoref{tab:sample_individual_attacks}) show that, while there are some individual attacks achieving higher performance compared to the ensembling for each dataset and counterfactual generation mechanism, there is no special attack that performs best for all. 
For instance, while distance-based attacks perform better when perturbation-based mechanisms like dice-gradient or \texttt{SCFE} are used, in the case of using instance-based mechanisms like \texttt{NICE} or \texttt{dice-kdtree}, neighbourhood-based and GAN-based attacks are more effective.
The ensembling benefits from the most effective attacks in the no-box setting, where no information is provided about the counterfactual generation mechanism or training data distribution.
Detailed results of individual attacks are presented in~\autoref{app:ind-attacks}.
\begin{table*}[ht]
\centering
\begin{tabular}{lllllll}
\hline
\textbf{Dataset} &\textbf{Attack} & \textbf{CF} & \textbf{ROC AUC} & \textbf{$\mathbf{TPR@FPR.01}$} & \textbf{$\mathbf{TPR@FPR.1}$} & \textbf{PR AUC} \\
\hline
\rowcolor{lightgray}
Compas & $GEN\_{lra}\_{K=20}$ & NICE & 0.621 & 0.006 & 0.180 & 0.595 \\
&$GEN\_{lra}\_{K=50}$ &  & 0.646 & 0.044 & 0.282 & 0.650 \\
\hline
\rowcolor{lightgray}
Heloc &$GEN\_{lra}\_{K=10}$ & Dice\_KDtree & 0.550 & 0.040 & 0.204 & 0.577 \\
 &$GEN\_{lra}\_{K=20}$ &  & 0.544 & 0.034 & 0.194 & 0.569 \\
\hline
\rowcolor{lightgray}
Acs\_income&$DCR\_{Diff\_{L_2}}$ & Dice\_Gradient & 0.504 & 0.013 & 0.114 & 0.513 \\
&$DCR\_{L_2}$ &  & 0.484 & 0.006 & 0.095 & 0.489 \\
\hline
\rowcolor{lightgray}
Adult &$DCR\_{L_2}$ & SCFE & 0.546 & 0.019 & 0.152 & 0.552 \\
& $DOMIAS$ &  & 0.515 & 0.010 & 0.093 & 0.508 \\
\hline
\end{tabular}
\caption{Sample results of Individual attacks.}
\label{tab:sample_individual_attacks}
\end{table*}

\section{Conclusion}
\label{chap:conclusion}
In this work, we evaluate the effectiveness of MIAs originally designed for synthetic data when applied to counterfactual explanations.
We have shown that successful MIAs are possible even without any access to the model or knowledge of the counterfactual generation mechanism.

Our results demonstrate that even without access to the model, successful MIAs against counterfactuals are possible, suggesting that practitioners should apply suitable privacy measures when generating counterfactuals. In particular, ensemble MIA works in a no-box setting, which is a weaker adversary model, while achieving better performance than the existing attacks targeting counterfactuals with query access to the model. For instance, we achieve an improvement of 26\% in Compas (dice-kdtree) which is significant enough to raise attention to the privacy issue of releasing counterfactuals without privacy measures, even when no further information or access is provided.
Finally, our results also highlight that the attack is more effective on smaller datasets and on more realistic counterfactuals. 
Moreover, adopting this no-box attack setting enables auditors to assess privacy leakage in deployed models without requiring any access to the models themselves.

To prevent no-box MIAs, the model providers might consider using differential privacy~\cite{dwork2014algorithmic} in counterfactual generation mechanisms~\cite{nelson2022privacy,pentyala2023privacy,huang2023accurate}.
Another defense technique is to return prototypic instances for each class, instead of generating query-specific counterfactuals. 
These techniques, by reducing the counterfactual set size available to the adversary, will limit the potential for performing successful MIAs.  
One limitation of our work is that even with the improvement of the membership inference attack performance using the no-box attack, which is a stronger attack with weaker access, for some datasets and counterfactuals, the performance of the attack is just slightly higher than a random guess. 
This low performance of the attack should not be mistaken for guaranteed privacy.
Privacy leakage can still exist, even when membership inference attacks fail.
To guarantee the privacy of data, practitioners should always consider implementing reliable countermeasures when generating and releasing counterfactuals.

\FloatBarrier

\begin{acks}
The authors acknowledge the support of the Digital Research Alliance of Canada (alliancecan.ca) and the advanced research computing resources made available through its national computing infrastructure. 
Héber H. Arcolezi is supported by the French National Research Agency (ANR) research grants (ANR-24-CE23-6239, ANR-23-IACL-0006). 
Ulrich Aïvodji is supported by the Fonds de recherche du Québec – Nature et technologies (FRQNT) Team Research Project grant (327090) and Natural Sciences and Engineering Research Council of Canada (NSERC) Discovery grant (RGPIN-2022-04006). 
Sébastien Gambs is supported by the Canada Research Chair in Privacy-preserving and Ethical Analysis of Big Data, FRQNT Team Project grant (327090) and NSERC Discovery grant (RGPIN-2022-05031).
\end{acks}

\bibliographystyle{ACM-Reference-Format}
\bibliography{sample-base}

\appendix
\section{Distribution comparison between real and counterfactual datasets} 
\label{app:distribution-comparison}
Comparing the distribution between the original and synthetic (Counterfactuals) datasets shows that the more these two distributions match, the more effectively the attack performs. 

\begin{figure}[ht]
\centering
\begin{tabular}{c c}

\begin{minipage}{0.45\textwidth}
    \centering
    \includegraphics[width=0.8\linewidth]{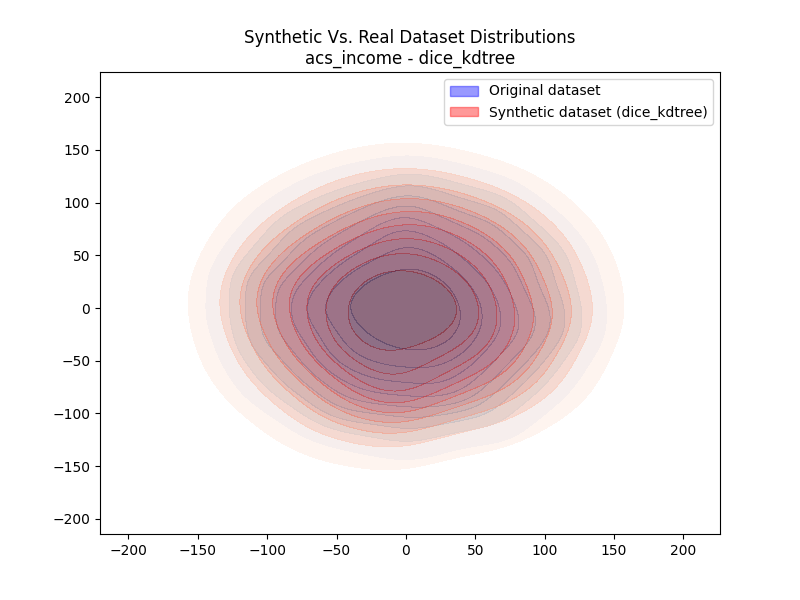}
    \\[-2pt]
    {\small dice\_kdtree}
\end{minipage}
&
\begin{minipage}{0.45\textwidth}
    \centering
    \includegraphics[width=0.8\linewidth]{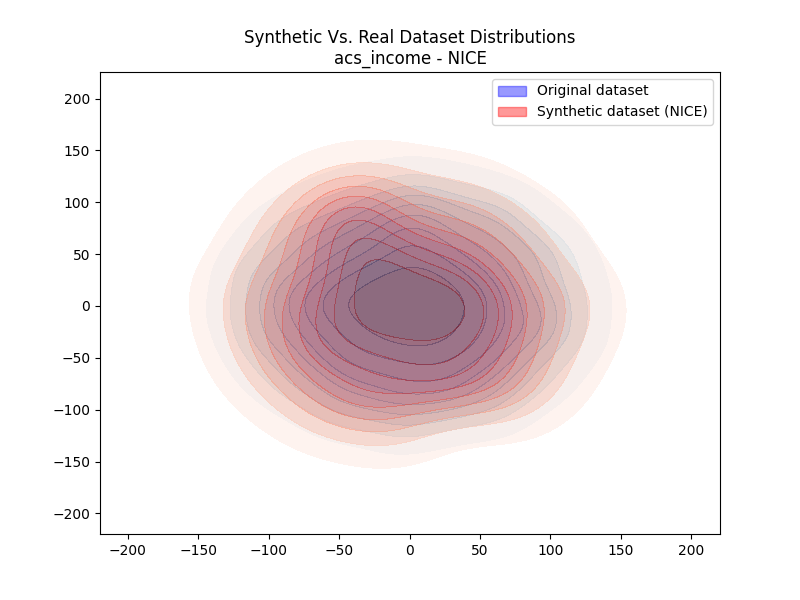}
    \\[-2pt]
    {\small Nice}
\end{minipage}
\\[10pt]

\begin{minipage}{0.45\textwidth}
    \centering
    \includegraphics[width=0.8\linewidth]{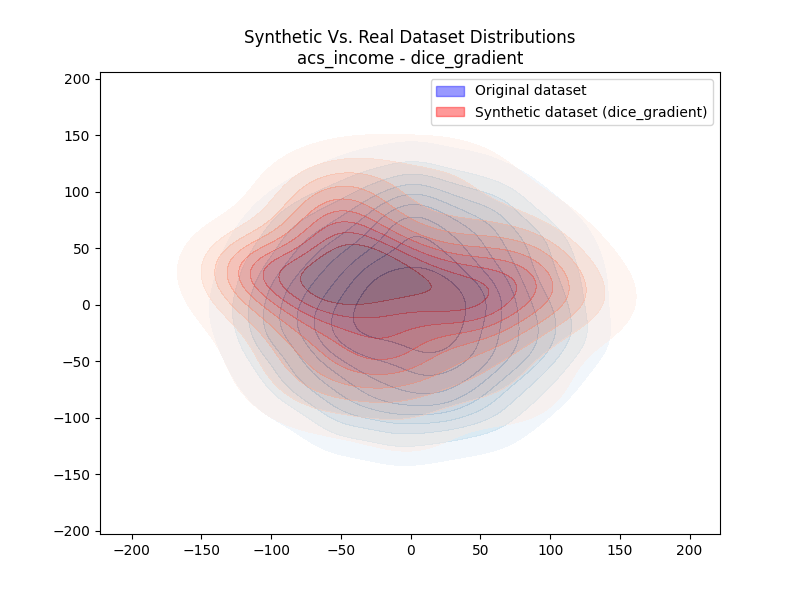}
    \\[-2pt]
    {\small dice\_gradient}
\end{minipage}
&
\begin{minipage}{0.45\textwidth}
    \centering
    \includegraphics[width=0.8\linewidth]{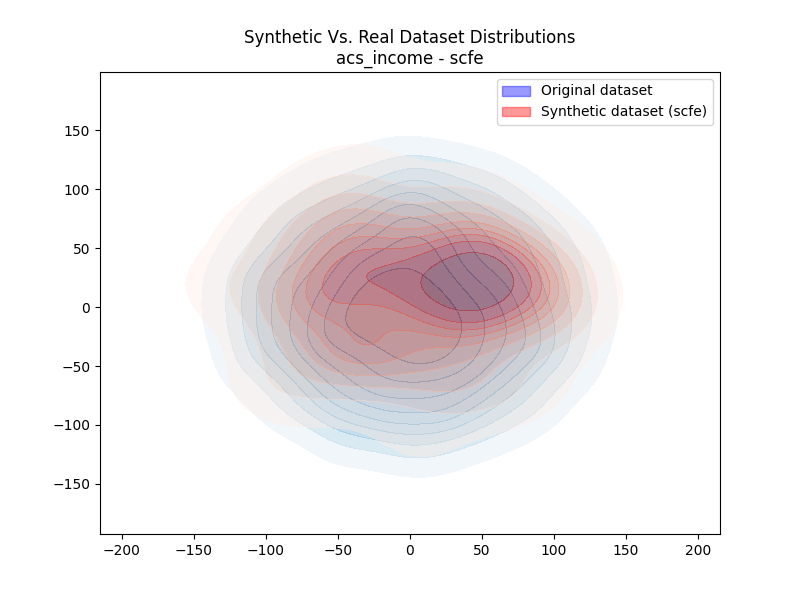}
    \\[-2pt]
    {\small scfe}
\end{minipage}
\\[10pt]




\end{tabular}

\caption{Acs\_income -- distribution\_comparison between real dataset and counterfactual sets used to perform no-box ensemble attack.}
\label{fig:acs-dist-comp}
\end{figure}
\begin{figure}[ht]
\centering
\begin{tabular}{c c}

\begin{minipage}{0.45\textwidth}
    \centering
    \includegraphics[width=0.8\linewidth]{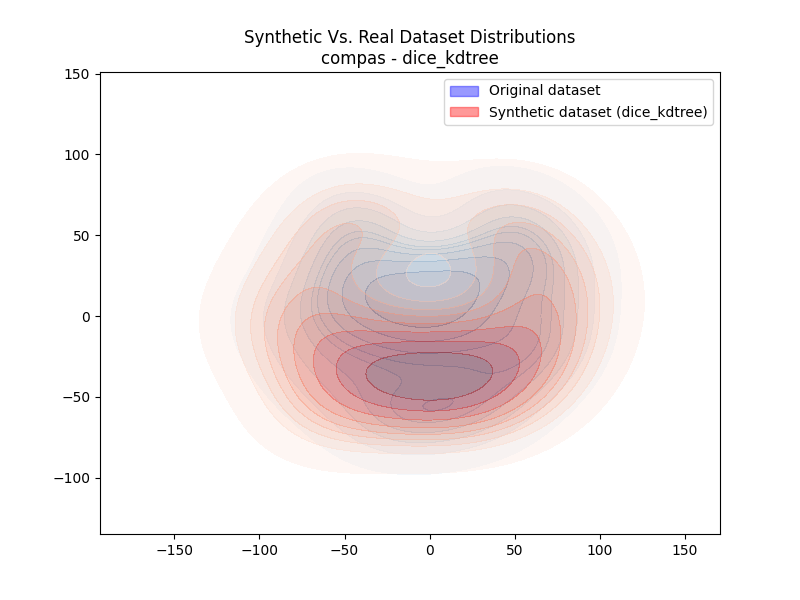}
    \\[-2pt]
    {\scriptsize dice\_kdtree}
\end{minipage}
&
\begin{minipage}{0.45\textwidth}
    \centering
    \includegraphics[width=0.8\linewidth]{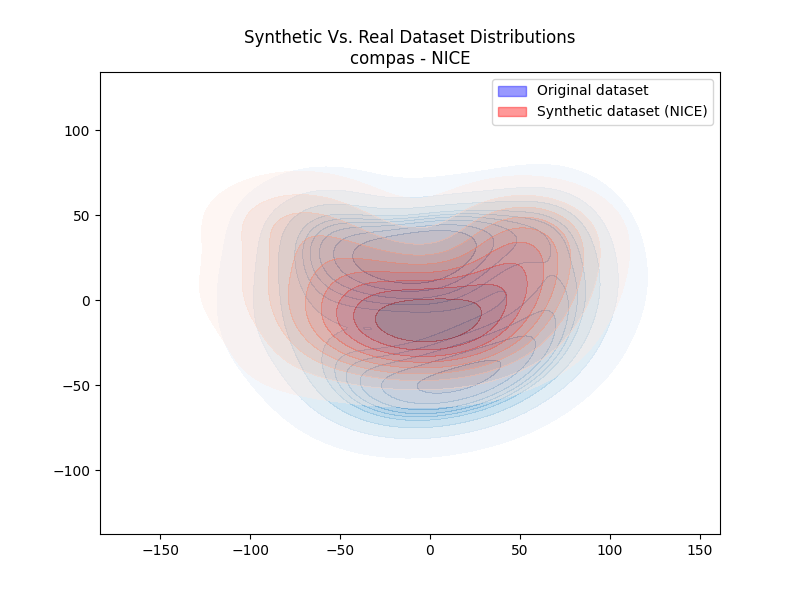}
    \\[-2pt]
    {\scriptsize Nice}
\end{minipage}
\\[10pt]

\begin{minipage}{0.45\textwidth}
    \centering
    \includegraphics[width=0.8\linewidth]{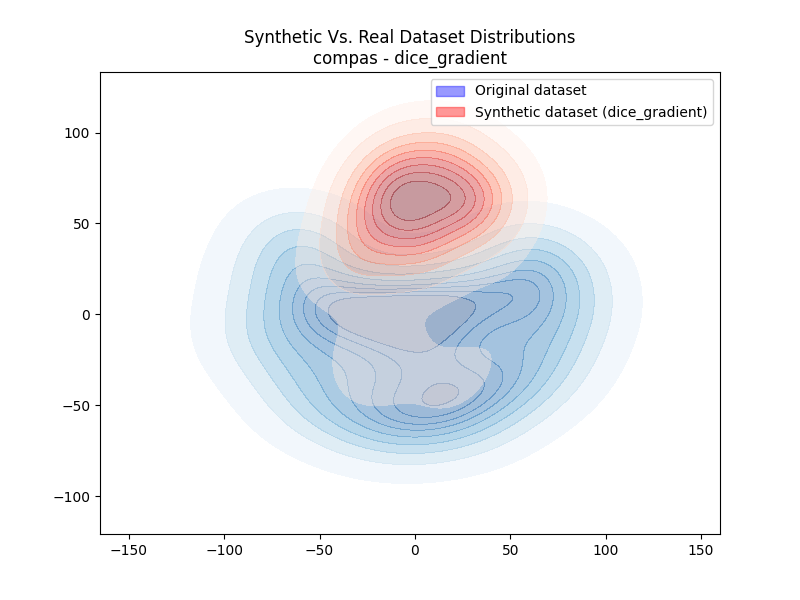}
    \\[-2pt]
    {\scriptsize dice\_gradient}
\end{minipage}
&
\begin{minipage}{0.45\textwidth}
    \centering
    \includegraphics[width=0.8\linewidth]{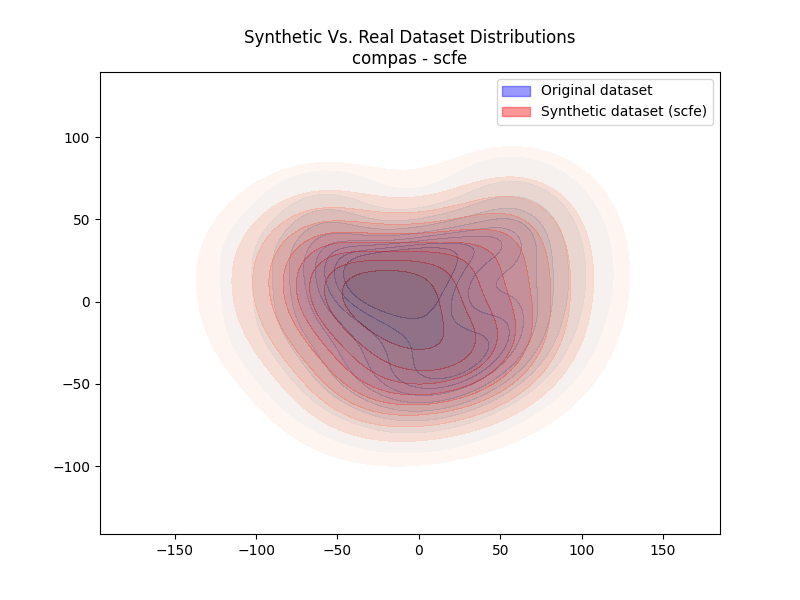}
    \\[-2pt]
    {\scriptsize scfe}
\end{minipage}
\\[10pt]




\end{tabular}

\caption{Compas -- distribution\_comparison between real dataset and counterfactual sets used to perform no-box ensemble attack.}
\label{fig:compas-dist-comp}
\end{figure}
\begin{figure}[ht]
\centering
\begin{tabular}{c c}

\begin{minipage}{0.45\textwidth}
    \centering
    \includegraphics[width=0.8\linewidth]{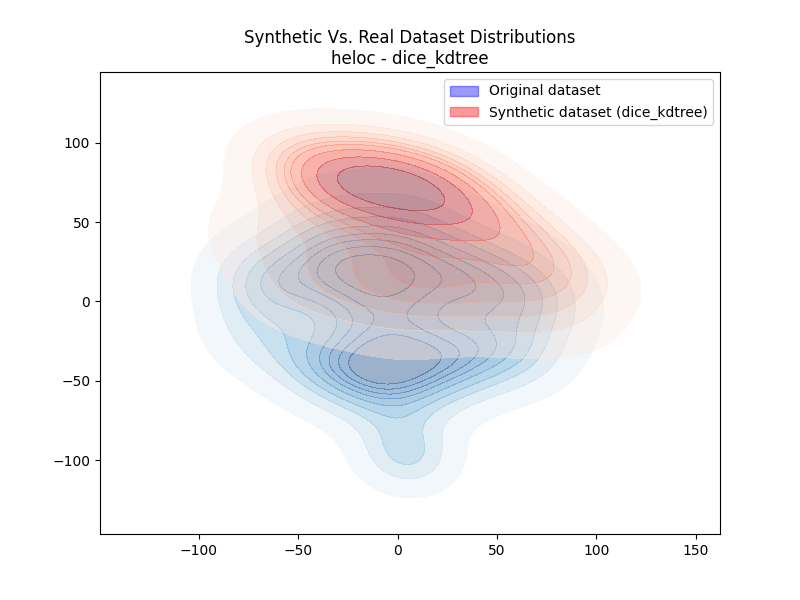}
    \\[-2pt]
    {\small dice\_kdtree}
\end{minipage}
&
\begin{minipage}{0.45\textwidth}
    \centering
    \includegraphics[width=0.8\linewidth]{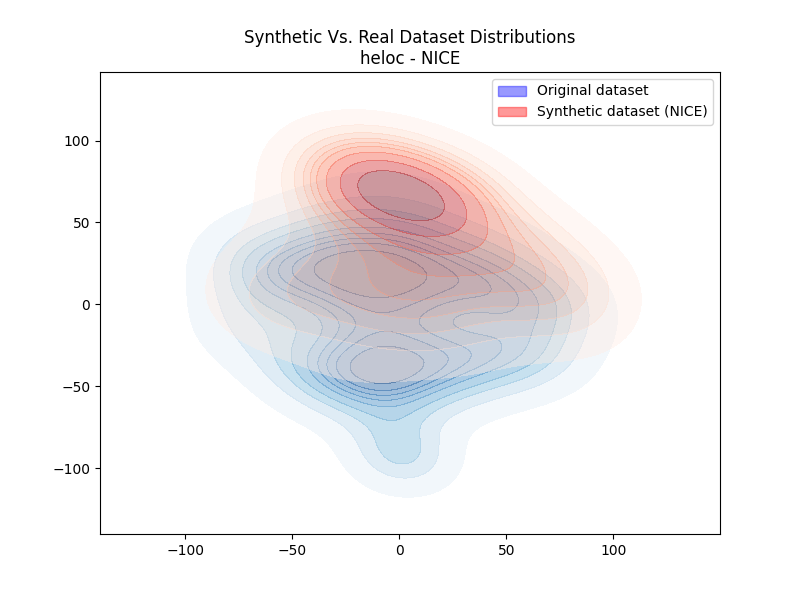}
    \\[-2pt]
    {\small Nice}
\end{minipage}
\\[10pt]

\begin{minipage}{0.45\textwidth}
    \centering
    \includegraphics[width=0.8\linewidth]{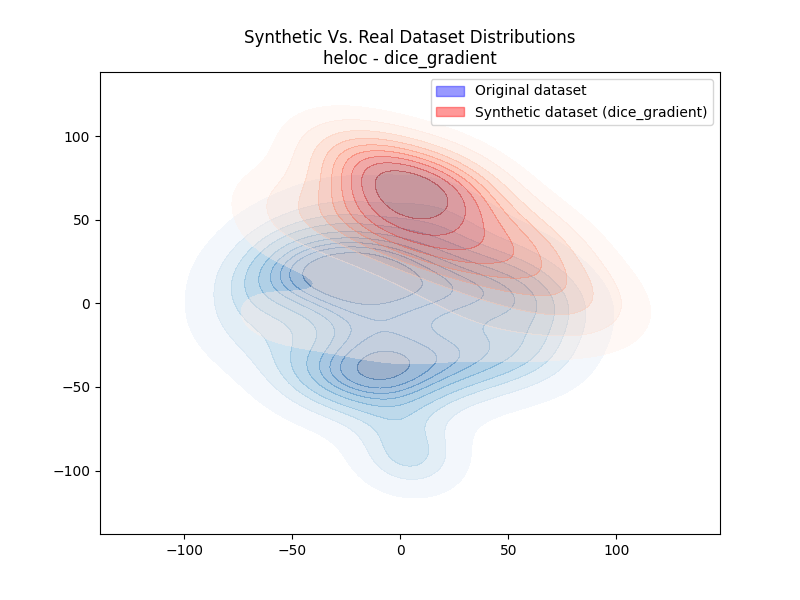}
    \\[-2pt]
    {\small dice\_gradient}
\end{minipage}
&
\begin{minipage}{0.45\textwidth}
    \centering
    \includegraphics[width=0.8\linewidth]{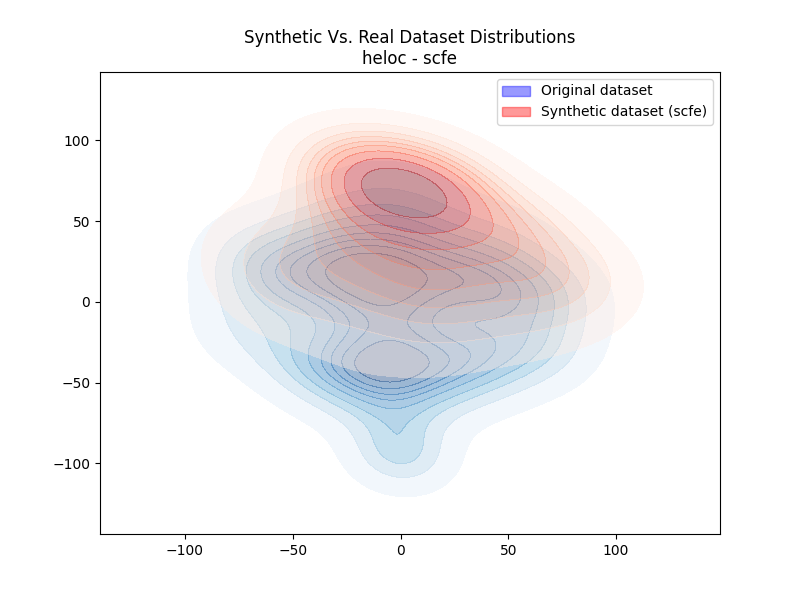}
    \\[-2pt]
    {\small scfe}
\end{minipage}
\\[10pt]




\end{tabular}

\caption{heloc -- distribution\_comparison between real dataset and counterfactual sets used to perform no-box ensemble attack.}
\label{fig:heloc-dist-comp}
\end{figure}
\begin{figure}[ht]
\centering
\begin{tabular}{c c}

\begin{minipage}{0.45\textwidth}
    \centering
    \includegraphics[width=0.8\linewidth]{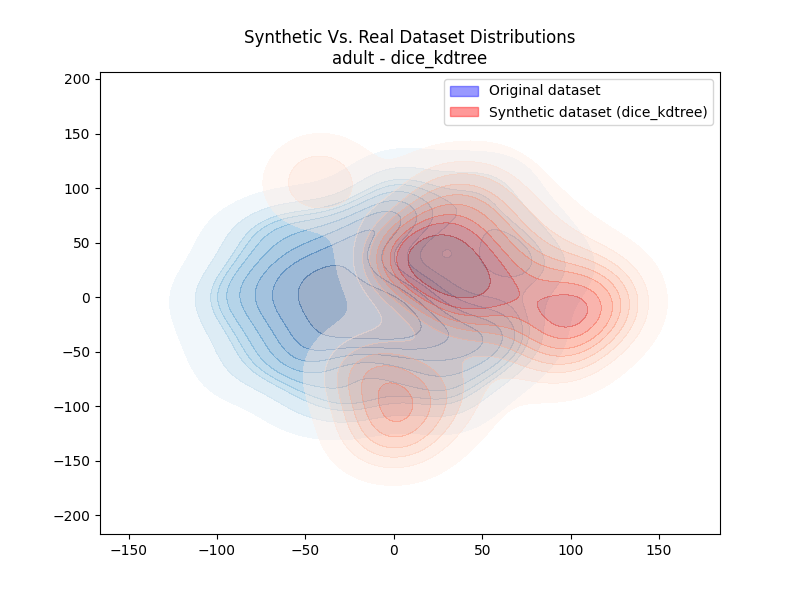}
    \\[-2pt]
    {\small dice\_kdtree}
\end{minipage}
&
\begin{minipage}{0.45\textwidth}
    \centering
    \includegraphics[width=0.8\linewidth]{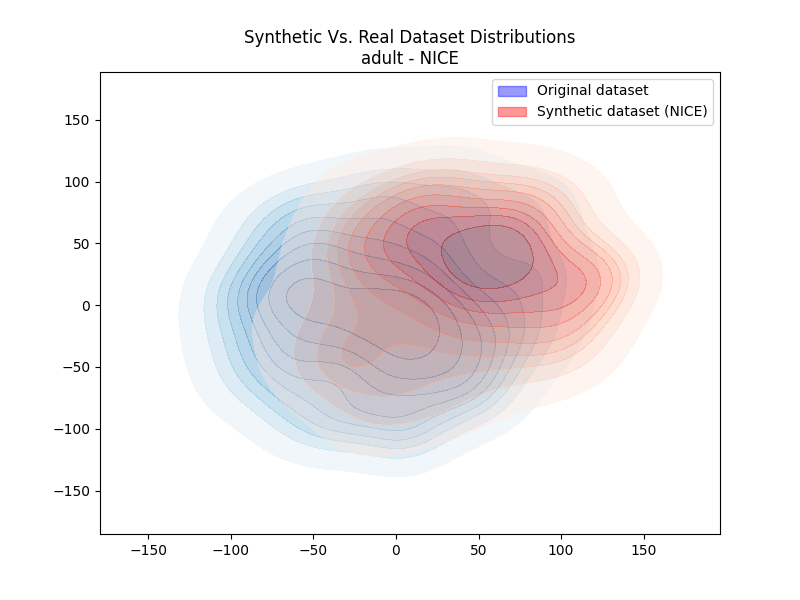}
    \\[-2pt]
    {\small Nice}
\end{minipage}
\\[10pt]

\begin{minipage}{0.45\textwidth}
    \centering
    \includegraphics[width=0.8\linewidth]{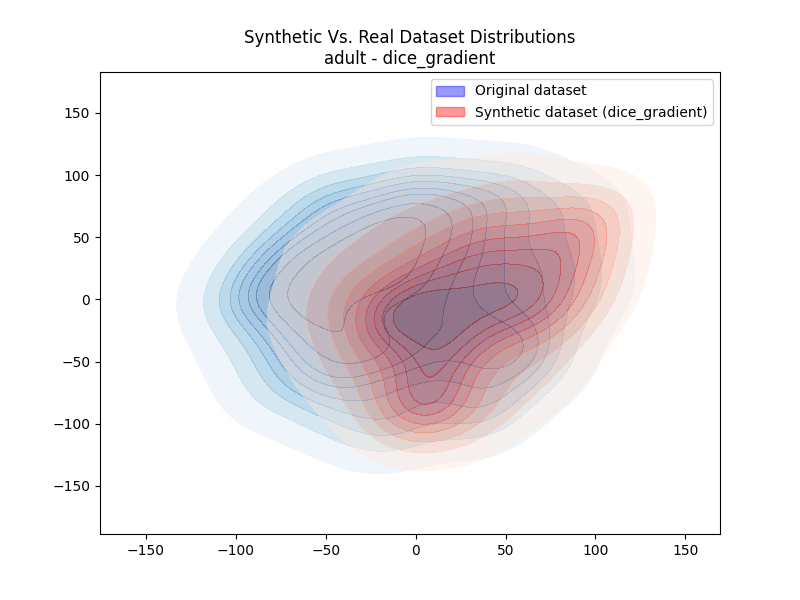}
    \\[-2pt]
    {\small dice\_gradient}
\end{minipage}
&
\begin{minipage}{0.45\textwidth}
    \centering
    \includegraphics[width=0.8\linewidth]{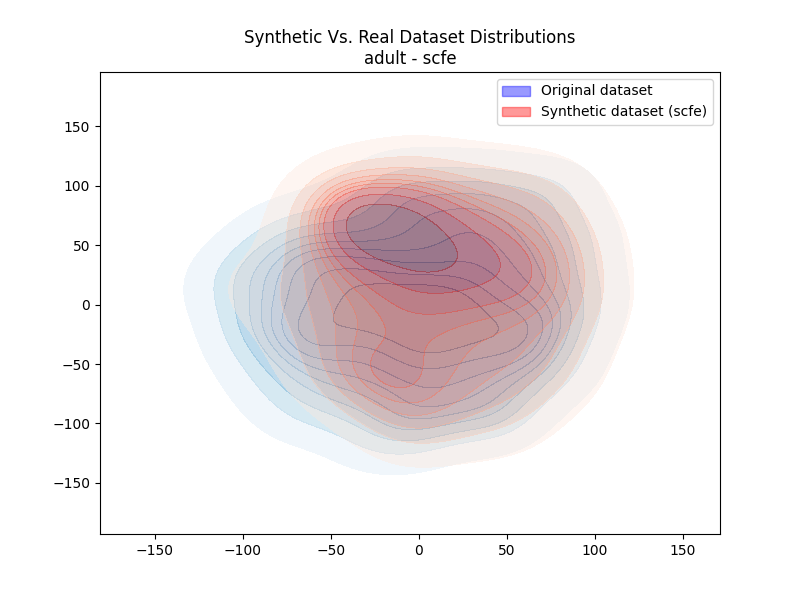}
    \\[-2pt]
    {\small scfe}
\end{minipage}
\\[10pt]




\end{tabular}

\caption{adult -- distribution\_comparison between real dataset and counterfactual sets used to perform no-box ensemble attack.}
\label{fig:adult-dist-comp}
\end{figure}

\FloatBarrier

\section{Individual attack results} \label{app:ind-attacks}
In this section, the results of individual attacks on various counterfactual techniques are presented. 
These results show that some individual attacks are working better than the ensemble attacks across datasets and counterfactual generation mechanisms, but since the best attack is not the same for all datasets, or all counterfactual generation techniques per dataset, in the no-box setting where the attacker has no information about the methods and data distributions, they cannot decide on the best attack, and ensembling improves the average attack performance.

\begin{table*}[ht]
\centering
\begin{tabular}{llllll}
\hline
\textbf{Attack} & \textbf{CF} & \textbf{ROC AUC} & \textbf{$\mathbf{TPR@FPR.01}$} & \textbf{$\mathbf{TPR@FPR.1}$} & \textbf{PR AUC} \\
\hline
$DCR\_{Diff\_{L_2}}$ & $NICE$ & 0.514 & 0.017 & 0.111 & 0.521 \\
\rowcolor{lightgray}
$DCR\_{L_2}$ &  & 0.497 & 0.019 & 0.109 & 0.509 \\
$DOMIAS$ &  & 0.503 & 0.018 & 0.111 & 0.511 \\
\rowcolor{lightgray}
$DPI\_{L_2}\_{K=10}$ &  & 0.509 & 0.011 & 0.105 & 0.507 \\
$DPI\_{L_2}\_{K=20}$ &  & 0.502 & 0.010 & 0.103 & 0.502 \\
\rowcolor{lightgray}
$GEN\_{LRA}\_{K=1}$ &  & 0.505 & 0.008 & 0.120 & 0.510 \\
$GEN\_{lra}\_{K=10}$ &  & 0.509 & 0.014 & 0.118 & 0.513 \\
\rowcolor{lightgray}
$GEN\_{lra}\_{K=20}$ &  & 0.508 & 0.010 & 0.117 & 0.512 \\
$GEN\_{lra}\_{K=50}$ &  & 0.506 & 0.008 & 0.108 & 0.508 \\
\rowcolor{lightgray}
$LOGAN$ &  & 0.509 & 0.017 & 0.102 & 0.511 \\
\hline
$DCR\_{Diff\_{L_2}}$ & $Dice\_Gradient$ & 0.504 & 0.013 & 0.114 & 0.513 \\
\rowcolor{lightgray}
$DCR\_{L_2}$ &  & 0.484 & 0.006 & 0.095 & 0.489 \\
$DOMIAS$ &  & 0.498 & 0.016 & 0.122 & 0.512 \\
\rowcolor{lightgray}
$DPI\_{L_2}\_{K=10}$ &  & 0.501 & 0.011 & 0.107 & 0.504 \\
$DPI\_{L_2}\_{K=20}$ &  & 0.500 & 0.011 & 0.111 & 0.503 \\
\rowcolor{lightgray}
$GEN\_{LRA}\_{K=1}$ &  & 0.496 & 0.015 & 0.106 & 0.509 \\
$GEN\_{lra}\_{K=10}$ &  & 0.499 & 0.014 & 0.119 & 0.511 \\
\rowcolor{lightgray}
$GEN\_{lra}\_{K=20}$ &  & 0.499 & 0.017 & 0.112 & 0.511 \\
$GEN\_{lra}\_{K=50}$ &  & 0.502 & 0.016 & 0.113 & 0.513 \\
\rowcolor{lightgray}
$LOGAN$ &  & 0.506 & 0.016 & 0.099 & 0.510 \\
\hline
$DCR\_{Diff\_{L_2}}$ & $Dice\_KDTree$ & 0.516 & 0.027 & 0.137 & 0.538 \\
\rowcolor{lightgray}
$DCR\_{L_2}$ &  & 0.507 & 0.048 & 0.127 & 0.538 \\
$DOMIAS$ &  & 0.506 & 0.022 & 0.127 & 0.527 \\
\rowcolor{lightgray}
$DPI\_{L_2}\_{K=10}$ &  & 0.502 & 0.010 & 0.104 & 0.503 \\
$DPI\_{L_2}\_{K=20}$ &  & 0.498 & 0.010 & 0.103 & 0.501 \\
\rowcolor{lightgray}
$GEN\_{LRA}\_{K=1}$ &  & 0.510 & 0.015 & 0.136 & 0.524 \\
$GEN\_{lra}\_{K=10}$ &  & 0.506 & 0.018 & 0.129 & 0.520 \\
\rowcolor{lightgray}
$GEN\_{lra}\_{K=20}$ &  & 0.504 & 0.017 & 0.124 & 0.516 \\
$GEN\_{lra}\_{K=50}$ &  & 0.502 & 0.018 & 0.107 & 0.509 \\
\rowcolor{lightgray}
$LOGAN$ &  & 0.507 & 0.016 & 0.117 & 0.514 \\
\hline
$DCR\_{Diff\_{L_2}}$ & $SCFE$ & 0.514 & 0.018 & 0.116 & 0.519 \\
\rowcolor{lightgray}
$DCR\_{L_2}$ &  & 0.490 & 0.014 & 0.102 & 0.499 \\
$DOMIAS$ &  & 0.507 & 0.015 & 0.122 & 0.516 \\
\rowcolor{lightgray}
$DPI\_{L_2}\_{K=10}$ &  & 0.507 & 0.010 & 0.105 & 0.505 \\
$DPI\_{L_2}\_{K=20}$ &  & 0.501 & 0.010 & 0.105 & 0.503 \\
\rowcolor{lightgray}
$GEN\_{LRA}\_{K=1}$ &  & 0.502 & 0.012 & 0.123 & 0.510 \\
$GEN\_{lra}\_{K=10}$ &  & 0.506 & 0.014 & 0.109 & 0.512 \\
\rowcolor{lightgray}
$GEN\_{lra}\_{K=20}$ &  & 0.504 & 0.013 & 0.118 & 0.510 \\
$GEN\_{lra}\_{K=50}$ &  & 0.505 & 0.010 & 0.111 & 0.508 \\
\rowcolor{lightgray}
$LOGAN$ &  & 0.509 & 0.016 & 0.116 & 0.515 \\
\hline
\end{tabular}
\caption{Individual attack results for $Acs\_Income$}
\label{tab:acs_income_individual_attacks.}
\end{table*}

\begin{table*}[ht]
\centering
\begin{tabular}{llllll}
\hline
\textbf{Attack} & \textbf{CF} & \textbf{ROC AUC} & \textbf{$\mathbf{TPR@FPR.01}$} & \textbf{$\mathbf{TPR@FPR.1}$} & \textbf{PR AUC} \\
\hline
$DCR\_{Diff\_{L_2}}$ & $NICE$ & 0.497 & 0.016 & 0.106 & 0.509 \\
\rowcolor{lightgray}
$DCR\_{L_2}$ &  & 0.537 & 0.057 & 0.161 & 0.563 \\
$DOMIAS$ &  & 0.525 & 0.015 & 0.137 & 0.529 \\
\rowcolor{lightgray}
$DPI\_{L_2}\_{K=10}$ &  & 0.501 & 0.011 & 0.105 & 0.504 \\
$DPI\_{L_2}\_{K=20}$ &  & 0.506 & 0.011 & 0.109 & 0.507 \\
\rowcolor{lightgray}
$GEN\_{LRA}\_{K=1}$ &  & 0.496 & 0.013 & 0.090 & 0.494 \\
$GEN\_{lra}\_{K=10}$ &  & 0.499 & 0.012 & 0.087 & 0.494 \\
\rowcolor{lightgray}
$GEN\_{lra}\_{K=20}$ &  & 0.502 & 0.010 & 0.087 & 0.497 \\
$GEN\_{lra}\_{K=50}$ &  & 0.508 & 0.014 & 0.088 & 0.503 \\
\rowcolor{lightgray}
$LOGAN$ &  & 0.456 & 0.011 & 0.090 & 0.477 \\
\hline
$DCR\_{Diff\_{L_2}}$ & $Dice\_Gradient$ & 0.508 & 0.013 & 0.106 & 0.510 \\
\rowcolor{lightgray}
$DCR\_{L_2}$ &  & 0.544 & 0.030 & 0.150 & 0.555 \\
$DOMIAS$ &  & 0.527 & 0.013 & 0.127 & 0.528 \\
\rowcolor{lightgray}
$DPI\_{L_2}\_{K=10}$ &  & 0.515 & 0.011 & 0.111 & 0.513 \\
$DPI\_{L_2}\_{K=20}$ &  & 0.514 & 0.011 & 0.106 & 0.512 \\
\rowcolor{lightgray}
$GEN\_{LRA}\_{K=1}$ &  & 0.500 & 0.008 & 0.077 & 0.491 \\
$GEN\_{lra}\_{K=10}$ &  & 0.498 & 0.009 & 0.072 & 0.489 \\
\rowcolor{lightgray}
$GEN\_{lra}\_{K=20}$ &  & 0.496 & 0.007 & 0.076 & 0.489 \\
$GEN\_{lra}\_{K=50}$ &  & 0.501 & 0.008 & 0.078 & 0.493 \\
\rowcolor{lightgray}
$LOGAN$ &  & 0.460 & 0.013 & 0.093 & 0.481 \\
\hline
$DCR\_{Diff\_{L_2}}$ & $Dice\_KDTree$ & 0.500 & 0.027 & 0.130 & 0.526 \\
\rowcolor{lightgray}
$DCR\_{L_2}$ &  & 0.547 & 0.082 & 0.161 & 0.580 \\
$DOMIAS$ &  & 0.521 & 0.022 & 0.124 & 0.529 \\
\rowcolor{lightgray}
$DPI\_{L_2}\_{K=10}$ &  & 0.490 & 0.011 & 0.105 & 0.500 \\
$DPI\_{L_2}\_{K=20}$ &  & 0.495 & 0.011 & 0.112 & 0.504 \\
\rowcolor{lightgray}
$GEN\_{LRA}\_{K=1}$ &  & 0.494 & 0.018 & 0.099 & 0.502 \\
$GEN\_{lra}\_{K=10}$ &  & 0.495 & 0.015 & 0.095 & 0.500 \\
\rowcolor{lightgray}
$GEN\_{lra}\_{K=20}$ &  & 0.495 & 0.015 & 0.098 & 0.500 \\
$GEN\_{lra}\_{K=50}$ &  & 0.503 & 0.013 & 0.096 & 0.503 \\
\rowcolor{lightgray}
$LOGAN$ &  & 0.464 & 0.014 & 0.092 & 0.483 \\
\hline
$DCR\_{Diff\_{L_2}}$ & $SCFE$ & 0.477 & 0.008 & 0.080 & 0.480 \\
\rowcolor{lightgray}
$DCR\_{L_2}$ &  & 0.546 & 0.019 & 0.152 & 0.552 \\
$DOMIAS$ &  & 0.515 & 0.010 & 0.093 & 0.508 \\
\rowcolor{lightgray}
$DPI\_{L_2}\_{K=10}$ &  & 0.503 & 0.010 & 0.102 & 0.503 \\
$DPI\_{L_2}\_{K=20}$ &  & 0.511 & 0.011 & 0.105 & 0.507 \\
\rowcolor{lightgray}
$GEN\_{LRA}\_{K=1}$ &  & 0.451 & 0.007 & 0.073 & 0.466 \\
$GEN\_{lra}\_{K=10}$ &  & 0.465 & 0.009 & 0.082 & 0.475 \\
\rowcolor{lightgray}
$GEN\_{lra}\_{K=20}$ &  & 0.477 & 0.008 & 0.085 & 0.482 \\
$GEN\_{lra}\_{K=50}$ &  & 0.499 & 0.010 & 0.094 & 0.497 \\
\rowcolor{lightgray}
$LOGAN$ &  & 0.452 & 0.010 & 0.090 & 0.474 \\
\hline
\end{tabular}
\caption{Individual attack results for Adult.}
\label{tab:adult_individual_attacks}
\end{table*}

\begin{table*}[ht]
\centering
\begin{tabular}{llllll}
\hline
\textbf{Attack} & \textbf{CF} & \textbf{ROC AUC} & \textbf{$\mathbf{TPR@FPR.01}$} & \textbf{$\mathbf{TPR@FPR.1}$} & \textbf{PR AUC} \\
\hline
$DCR\_{Diff\_{L_2}}$ & $NICE$ & 0.607 & 0.013 & 0.146 & 0.579 \\
\rowcolor{lightgray}
$DCR\_{L_2}$ &  & 0.642 & 0.055 & 0.240 & 0.638 \\
$DOMIAS$ &  & 0.634 & 0.067 & 0.272 & 0.649 \\
\rowcolor{lightgray}
$DPI\_{L_2}\_{K=10}$ &  & 0.626 & 0.022 & 0.224 & 0.609 \\
$DPI\_{L_2}\_{K=20}$ &  & 0.641 & 0.036 & 0.292 & 0.638 \\
\rowcolor{lightgray}
$GEN\_{LRA}\_{K=1}$ &  & 0.585 & 0.012 & 0.086 & 0.542 \\
$GEN\_{lra}\_{K=10}$ &  & 0.605 & 0.007 & 0.139 & 0.571 \\
\rowcolor{lightgray}
$GEN\_{lra}\_{K=20}$ &  & 0.621 & 0.006 & 0.180 & 0.595 \\
$GEN\_{lra}\_{K=50}$ &  & 0.646 & 0.044 & 0.282 & 0.650 \\
\rowcolor{lightgray}
$LOGAN$ &  & 0.357 & 0.006 & 0.050 & 0.413 \\
\hline
$DCR\_{Diff\_{L_2}}$ & $Dice\_Gradient$ & 0.595 & 0.018 & 0.162 & 0.576 \\
\rowcolor{lightgray}
$DCR\_{L_2}$ &  & 0.621 & 0.051 & 0.206 & 0.619 \\
$DOMIAS$ &  & 0.571 & 0.006 & 0.098 & 0.537 \\
\rowcolor{lightgray}
$DPI\_{L_2}\_{K=10}$ &  & 0.572 & 0.026 & 0.178 & 0.566 \\
$DPI\_{L_2}\_{K=20}$ &  & 0.595 & 0.019 & 0.177 & 0.583 \\
\rowcolor{lightgray}
$GEN\_{LRA}\_{K=1}$ &  & 0.541 & 0.003 & 0.114 & 0.531 \\
$GEN\_{lra}\_{K=10}$ &  & 0.541 & 0.007 & 0.100 & 0.525 \\
\rowcolor{lightgray}
$GEN\_{lra}\_{K=20}$ &  & 0.521 & 0.007 & 0.086 & 0.510 \\
$GEN\_{lra}\_{K=50}$ &  & 0.519 & 0.008 & 0.060 & 0.505 \\
\rowcolor{lightgray}
$LOGAN$ &  & 0.418 & 0.005 & 0.052 & 0.440 \\
\hline
$DCR\_{Diff\_{L_2}}$ & $Dice\_KDTree$ & 0.640 & 0.010 & 0.302 & 0.641 \\
\rowcolor{lightgray}
$DCR\_{L_2}$ &  & 0.660 & 0.226 & 0.339 & 0.708 \\
$DOMIAS$ &  & 0.645 & 0.068 & 0.350 & 0.681 \\
\rowcolor{lightgray}
$DPI\_{L_2}\_{K=10}$ &  & 0.639 & 0.033 & 0.316 & 0.639 \\
$DPI\_{L_2}\_{K=20}$ &  & 0.652 & 0.045 & 0.322 & 0.657 \\
\rowcolor{lightgray}
$GEN\_{LRA}\_{K=1}$ &  & 0.613 & 0.008 & 0.160 & 0.576 \\
$GEN\_{lra}\_{K=10}$ &  & 0.640 & 0.002 & 0.268 & 0.618 \\
\rowcolor{lightgray}
$GEN\_{lra}\_{K=20}$ &  & 0.653 & 0.022 & 0.302 & 0.647 \\
$GEN\_{lra}\_{K=50}$ &  & 0.670 & 0.048 & 0.332 & 0.679 \\
\rowcolor{lightgray}
$LOGAN$ &  & 0.340 & 0.004 & 0.056 & 0.413 \\
\hline
$DCR\_{Diff\_{L_2}}$ & $SCFE$ & 0.587 & 0.005 & 0.082 & 0.531 \\
\rowcolor{lightgray}
$DCR\_{L_2}$ &  & 0.697 & 0.070 & 0.347 & 0.704 \\
$DOMIAS$ &  & 0.691 & 0.006 & 0.307 & 0.658 \\
\rowcolor{lightgray}
$DPI\_{L_2}\_{K=10}$ &  & 0.683 & 0.032 & 0.316 & 0.655 \\
$DPI\_{L_2}\_{K=20}$ &  & 0.699 & 0.056 & 0.372 & 0.690 \\
\rowcolor{lightgray}
$GEN\_{LRA}\_{K=1}$ &  & 0.508 & 0.007 & 0.053 & 0.483 \\
$GEN\_{lra}\_{K=10}$ &  & 0.603 & 0.008 & 0.110 & 0.559 \\
\rowcolor{lightgray}
$GEN\_{lra}\_{K=20}$ &  & 0.652 & 0.006 & 0.206 & 0.611 \\
$GEN\_{lra}\_{K=50}$ &  & 0.690 & 0.025 & 0.341 & 0.677 \\
\rowcolor{lightgray}
$LOGAN$ &  & 0.352 & 0.008 & 0.050 & 0.415 \\
\hline
\end{tabular}
\caption{Individual attack results for $Compas$.}
\label{tab:compas_individual_attacks}
\end{table*}

\begin{table*}[ht]
\centering
\begin{tabular}{llllll}
\hline
\textbf{Attack} & \textbf{CF} & \textbf{ROC AUC} & \textbf{$\mathbf{TPR@FPR.01}$} & \textbf{$\mathbf{TPR@FPR.1}$} & \textbf{PR AUC} \\
\hline
$DCR\_{Diff\_{L_2}}$ & $NICE$ & 0.512 & 0.032 & 0.131 & 0.529 \\
\rowcolor{lightgray}
$DCR\_{L_2}$ &  & 0.520 & 0.023 & 0.108 & 0.527 \\
$DOMIAS$ &  & 0.503 & 0.016 & 0.108 & 0.512 \\
\rowcolor{lightgray}
$DPI\_{L_2}\_{K=10}$ &  & 0.495 & 0.012 & 0.108 & 0.501 \\
$DPI\_{L_2}\_{K=20}$ &  & 0.491 & 0.011 & 0.105 & 0.498 \\
\rowcolor{lightgray}
$GEN\_{LRA}\_{K=1}$ &  & 0.515 & 0.021 & 0.139 & 0.527 \\
$GEN\_{lra}\_{K=10}$ &  & 0.514 & 0.016 & 0.126 & 0.523 \\
\rowcolor{lightgray}
$GEN\_{lra}\_{K=20}$ &  & 0.515 & 0.017 & 0.119 & 0.519 \\
$GEN\_{lra}\_{K=50}$ &  & 0.512 & 0.014 & 0.113 & 0.513 \\
\rowcolor{lightgray}
$LOGAN$ &  & 0.498 & 0.011 & 0.103 & 0.502 \\
\hline
$DCR\_{Diff\_{L_2}}$ & $Dice\_Gradient$ & 0.481 & 0.013 & 0.098 & 0.493 \\
\rowcolor{lightgray}
$DCR\_{L_2}$ &  & 0.492 & 0.008 & 0.103 & 0.497 \\
$DOMIAS$ &  & 0.483 & 0.011 & 0.094 & 0.491 \\
\rowcolor{lightgray}
$DPI\_{L_2}\_{K=10}$ &  & 0.475 & 0.009 & 0.085 & 0.485 \\
$DPI\_{L_2}\_{K=20}$ &  & 0.480 & 0.010 & 0.093 & 0.488 \\
\rowcolor{lightgray}
$GEN\_{LRA}\_{K=1}$ &  & 0.505 & 0.011 & 0.102 & 0.505 \\
$GEN\_{lra}\_{K=10}$ &  & 0.498 & 0.007 & 0.102 & 0.501 \\
\rowcolor{lightgray}
$GEN\_{lra}\_{K=20}$ &  & 0.496 & 0.010 & 0.101 & 0.498 \\
$GEN\_{lra}\_{K=50}$ &  & 0.495 & 0.006 & 0.094 & 0.494 \\
\rowcolor{lightgray}
$LOGAN$ &  & 0.502 & 0.013 & 0.103 & 0.504 \\
\hline
$DCR\_{Diff\_{L_2}}$ & $Dice\_KDTree$ & 0.539 & 0.150 & 0.213 & 0.611 \\
\rowcolor{lightgray}
$DCR\_{L_2}$ &  & 0.546 & 0.148 & 0.205 & 0.612 \\
$DOMIAS$ &  & 0.542 & 0.117 & 0.206 & 0.602 \\
\rowcolor{lightgray}
$DPI\_{L_2}\_{K=10}$ &  & 0.492 & 0.011 & 0.106 & 0.499 \\
$DPI\_{L_2}\_{K=20}$ &  & 0.488 & 0.012 & 0.111 & 0.499 \\
\rowcolor{lightgray}
$GEN\_{LRA}\_{K=1}$ &  & 0.543 & 0.079 & 0.212 & 0.597 \\
$GEN\_{lra}\_{K=10}$ &  & 0.550 & 0.040 & 0.204 & 0.577 \\
\rowcolor{lightgray}
$GEN\_{lra}\_{K=20}$ &  & 0.544 & 0.034 & 0.194 & 0.569 \\
$GEN\_{lra}\_{K=50}$ &  & 0.532 & 0.027 & 0.176 & 0.556 \\
\rowcolor{lightgray}
$LOGAN$ &  & 0.497 & 0.015 & 0.103 & 0.500 \\
\hline
$DCR\_{Diff\_{L_2}}$ & $SCFE$ & 0.507 & 0.011 & 0.093 & 0.504 \\
\rowcolor{lightgray}
$DCR\_{L_2}$ &  & 0.509 & 0.013 & 0.106 & 0.507 \\
$DOMIAS$ &  & 0.505 & 0.009 & 0.100 & 0.504 \\
\rowcolor{lightgray}
$DPI\_{L_2}\_{K=10}$ &  & 0.502 & 0.009 & 0.093 & 0.500 \\
$DPI\_{L_2}\_{K=20}$ &  & 0.494 & 0.010 & 0.098 & 0.499 \\
\rowcolor{lightgray}
$GEN\_{LRA}\_{K=1}$ &  & 0.510 & 0.010 & 0.100 & 0.509 \\
$GEN\_{lra}\_{K=10}$ &  & 0.505 & 0.012 & 0.114 & 0.508 \\
\rowcolor{lightgray}
$GEN\_{lra}\_{K=20}$ &  & 0.503 & 0.014 & 0.111 & 0.507 \\
$GEN\_{lra}\_{K=50}$ &  & 0.507 & 0.012 & 0.114 & 0.510 \\
\rowcolor{lightgray}
$LOGAN$ &  & 0.495 & 0.010 & 0.101 & 0.499 \\
\hline
\end{tabular}
\caption{Individual attack results for $Heloc$.}
\label{tab:heloc_individual_attacks}
\end{table*}

\FloatBarrier

\section{Ablation study results} \label{app:ablation-atudies}
To evaluate the effectiveness of the no-box attack when the size of the synthetic data is small, we performed an ablation study over various synthetic set-attack set sizes. 
Two groups of experiments have been executed: First, by keeping the size fixed of the synthetic dataset (counterfactuals), we evaluated the effectiveness of the no-box attack on varying attack set sizes, changing from 200 to 2000 instances (except for the Compas dataset that is the smallest dataset in our experiment, both synthetic and attack set sizes for the experiments regarding this dataset are small as well). 
This study shows that for small datasets like Compas and Heloc, having an attack dataset as small as 200 instances can result in the best attack performance for most counterfactual methods. 
For larger datasets like $acs\_income$ and adult, attack set sizes of 1000 and 2000 can result in more accurate attacks.

\begin{table*}[ht]
\centering
\begin{tabular}{llllll}
\hline
\textbf{CF} & \textbf{Attack Set Size} & \textbf{ROC AUC} & \textbf{$\mathbf{TPR@FPR.01}$} & \textbf{$\mathbf{TPR@FPR.1}$} & \textbf{PR AUC} \\
\hline
$NICE$ & 200 & 0.490 & 0.015 & 0.087 & 0.507 \\
\rowcolor{lightgray}
 & 400 & 0.504 & 0.022 & 0.102 & 0.513 \\
 & 1000 & 0.495 & 0.013 & 0.096 & 0.501 \\
\rowcolor{lightgray}
 & 2000 & 0.509 & 0.011 & 0.107 & 0.510 \\
\hline
$Dice\_Gradient$ & 200 & 0.500 & 0.023 & 0.101 & 0.523 \\
\rowcolor{lightgray}
 & 400 & 0.482 & 0.010 & 0.078 & 0.497 \\
 & 1000 & 0.495 & 0.017 & 0.122 & 0.512 \\
\rowcolor{lightgray}
 & 2000 & 0.506 & 0.016 & 0.117 & 0.512 \\
\hline
$Dice\_K-DTree$ & 200 & 0.507 & 0.012 & 0.107 & 0.521 \\
\rowcolor{lightgray}
 & 400 & 0.503 & 0.030 & 0.117 & 0.523 \\
 & 1000 & 0.507 & 0.026 & 0.118 & 0.519 \\
\rowcolor{lightgray}
 & 2000 & 0.510 & 0.024 & 0.128 & 0.523 \\
\hline
$SCFE$ & 200 & 0.479 & 0.008 & 0.098 & 0.502 \\
\rowcolor{lightgray}
 & 400 & 0.501 & 0.021 & 0.111 & 0.514 \\
 & 1000 & 0.495 & 0.015 & 0.109 & 0.505 \\
\rowcolor{lightgray}
 & 2000 & 0.510 & 0.014 & 0.108 & 0.509 \\
\hline
\end{tabular}
\caption{Acs\_Income - Fixed Synth Size 10000.}
\label{tab:acs_income_10000}
\end{table*}

\begin{table*}[ht]
\centering
\begin{tabular}{llllll}
\hline
\textbf{CF} & \textbf{Attack Set Size} & \textbf{ROC AUC} & \textbf{$\mathbf{TPR@FPR.01}$} & \textbf{$\mathbf{TPR@FPR.1}$} & \textbf{PR AUC} \\
\hline
$NICE$ & 200 & 0.510 & 0.010 & 0.106 & 0.523 \\
\rowcolor{lightgray}
 & 400 & 0.499 & 0.008 & 0.080 & 0.501 \\
 & 1000 & 0.510 & 0.008 & 0.082 & 0.505 \\
\rowcolor{lightgray}
 & 2000 & 0.506 & 0.008 & 0.088 & 0.504 \\
\hline
$Dice\_Gradient$ & 200 & 0.522 & 0.007 & 0.130 & 0.537 \\
\rowcolor{lightgray}
 & 400 & 0.502 & 0.006 & 0.089 & 0.507 \\
 & 1000 & 0.515 & 0.008 & 0.102 & 0.514 \\
\rowcolor{lightgray}
 & 2000 & 0.515 & 0.008 & 0.101 & 0.512 \\
\hline
$Dice\_K-DTree$ & 200 & 0.525 & 0.014 & 0.126 & 0.536 \\
\rowcolor{lightgray}
 & 400 & 0.517 & 0.013 & 0.117 & 0.522 \\
 & 1000 & 0.525 & 0.012 & 0.109 & 0.520 \\
\rowcolor{lightgray}
 & 2000 & 0.506 & 0.013 & 0.112 & 0.509 \\
\hline
$SCFE$ & 200 & 0.478 & 0.010 & 0.103 & 0.495 \\
\rowcolor{lightgray}
 & 400 & 0.491 & 0.012 & 0.081 & 0.489 \\
 & 1000 & 0.495 & 0.010 & 0.096 & 0.497 \\
\rowcolor{lightgray}
 & 2000 & 0.488 & 0.008 & 0.083 & 0.488 \\
\hline
\end{tabular}
\caption{Adult - Fixed Synth Size 10000.}
\label{tab:adult_10000}
\end{table*}

\begin{table*}[ht]
\centering
\begin{tabular}{llllll}
\hline
\textbf{CF} & \textbf{Attack Set Size} & \textbf{ROC AUC} & \textbf{$\mathbf{TPR@FPR.01}$} & \textbf{$\mathbf{TPR@FPR.1}$} & \textbf{PR AUC} \\
\hline
$NICE$ & 200 & 0.615 & 0.010 & 0.164 & 0.604 \\
\rowcolor{lightgray}
 & 400 & 0.610 & 0.022 & 0.221 & 0.610 \\
 & 1000 & 0.616 & 0.011 & 0.177 & 0.601 \\
\hline
\rowcolor{lightgray}
$Dice\_Gradient$ & 200 & 0.609 & 0.033 & 0.213 & 0.607 \\
 & 400 & 0.607 & 0.018 & 0.218 & 0.600 \\
\rowcolor{lightgray}
 & 1000 & 0.603 & 0.024 & 0.204 & 0.593 \\
\hline
$Dice\_K-DTree$ & 200 & 0.685 & 0.047 & 0.351 & 0.667 \\
\rowcolor{lightgray}
 & 400 & 0.617 & 0.014 & 0.235 & 0.618 \\
 & 1000 & 0.618 & 0.008 & 0.180 & 0.606 \\
\hline
\rowcolor{lightgray}
$SCFE$ & 200 & 0.680 & 0.052 & 0.302 & 0.665 \\
 & 400 & 0.674 & 0.021 & 0.301 & 0.657 \\
\rowcolor{lightgray}
 & 1000 & 0.694 & 0.026 & 0.336 & 0.684 \\
\hline
\end{tabular}
\caption{Compas - Fixed Synth Size 2000.}
\label{tab:compas_2000}
\end{table*}

\begin{table*}[ht]
\centering
\begin{tabular}{llllll}
\hline
\textbf{CF} & \textbf{Attack Set Size} & \textbf{ROC AUC} & \textbf{$\mathbf{TPR@FPR.01}$} & \textbf{$\mathbf{TPR@FPR.1}$} & \textbf{PR AUC} \\
\hline
$NICE$ & 200 & 0.509 & 0.031 & 0.162 & 0.539 \\
\rowcolor{lightgray}
 & 400 & 0.481 & 0.016 & 0.090 & 0.492 \\
 & 1000 & 0.506 & 0.020 & 0.120 & 0.517 \\
\rowcolor{lightgray}
 & 2000 & 0.513 & 0.021 & 0.126 & 0.522 \\
\hline
$Dice\_Gradient$ & 200 & 0.474 & 0.036 & 0.130 & 0.521 \\
\rowcolor{lightgray}
 & 400 & 0.497 & 0.017 & 0.099 & 0.509 \\
 & 1000 & 0.502 & 0.010 & 0.123 & 0.509 \\
\rowcolor{lightgray}
 & 2000 & 0.486 & 0.009 & 0.091 & 0.494 \\
\hline
$Dice\_K-DTree$ & 200 & 0.549 & 0.146 & 0.212 & 0.596 \\
\rowcolor{lightgray}
 & 400 & 0.506 & 0.086 & 0.168 & 0.548 \\
 & 1000 & 0.531 & 0.067 & 0.165 & 0.563 \\
\rowcolor{lightgray}
 & 2000 & 0.537 & 0.060 & 0.175 & 0.564 \\
\hline
$SCFE$ & 200 & 0.515 & 0.039 & 0.141 & 0.542 \\
\rowcolor{lightgray}
 & 400 & 0.476 & 0.026 & 0.086 & 0.500 \\
 & 1000 & 0.507 & 0.012 & 0.100 & 0.508 \\
\rowcolor{lightgray}
 & 2000 & 0.496 & 0.008 & 0.097 & 0.498 \\
\hline
\end{tabular}
\caption{Heloc - Fixed Synth Size 10000.}
\label{tab:heloc_10000}
\end{table*}

\FloatBarrier
The other ablation study we performed fixed the attack set size and varied the synthetic data size to evaluate how many counterfactuals are needed to perform a successful no-box attack. 
These results show that while for larger datasets like $Acs\_income$ and Adult, the synthetic set size of 5000 to 10000 instances maximizes the attack performance, for smaller datasets like Compas and Heloc, a synthetic set size of 500 to 1000 instances is enough for the adversary user to perform a successful attack.

\begin{table*}[ht]
\centering
\begin{tabular}{llllll}
\hline
\textbf{CF} & \textbf{Synth Size} &  \textbf{ROC AUC} & \textbf{$\mathbf{TPR@FPR.01}$} & \textbf{$\mathbf{TPR@FPR.1}$} & \textbf{PR AUC} \\
\hline
$NICE$ & 500& 0.507 & 0.013 & 0.101 & 0.506 \\
\rowcolor{lightgray}
 & 1000& 0.508 & 0.013 & 0.101 & 0.508 \\
 & 2000& 0.503 & 0.013 & 0.103 & 0.506 \\
\rowcolor{lightgray}
 & 5000& 0.508 & 0.013 & 0.100 & 0.506 \\
 & 10000& 0.509 & 0.011 & 0.107 & 0.510 \\
\hline
\rowcolor{lightgray}
$Dice\_Gradient$ & 500& 0.505 & 0.011 & 0.103 & 0.505 \\
 & 1000& 0.503 & 0.016 & 0.110 & 0.508 \\
\rowcolor{lightgray}
 & 2000& 0.502 & 0.016 & 0.113 & 0.510 \\
 & 5000& 0.505 & 0.017 & 0.120 & 0.512 \\
\rowcolor{lightgray}
 & 10000& 0.506 & 0.016 & 0.117 & 0.512 \\
\hline
$Dice\_K-DTree$ & 500& 0.505 & 0.013 & 0.105 & 0.505 \\
\rowcolor{lightgray}
 & 1000& 0.505 & 0.016 & 0.111 & 0.508 \\
 & 2000& 0.507 & 0.014 & 0.112 & 0.510 \\
\rowcolor{lightgray}
 & 5000& 0.511 & 0.022 & 0.113 & 0.516 \\
 & 10000& 0.510 & 0.024 & 0.128 & 0.523 \\
\hline
\rowcolor{lightgray}
$SCFE$ & 500& 0.510 & 0.014 & 0.122 & 0.511 \\
 & 1000& 0.503 & 0.013 & 0.105 & 0.506 \\
\rowcolor{lightgray}
 & 2000& 0.504 & 0.016 & 0.106 & 0.508 \\
 & 5000& 0.511 & 0.015 & 0.104 & 0.510 \\
\rowcolor{lightgray}
 & 10000& 0.510 & 0.014 & 0.108 & 0.509 \\
\hline
\end{tabular}
\caption{$Acs\_Income$ - Fixed Attack Set Size 1000.}
\label{tab:Acs_Income_attack_1000}
\end{table*}

\begin{table*}[ht]
\centering
\begin{tabular}{llllll}
\hline
\textbf{CF} & \textbf{Synth Size} &  \textbf{ROC AUC} & \textbf{$\mathbf{TPR@FPR.01}$} & \textbf{$\mathbf{TPR@FPR.1}$} & \textbf{PR AUC} \\
\hline
$NICE$ & 500& 0.516 & 0.008 & 0.100 & 0.514 \\
\rowcolor{lightgray}
 & 1000& 0.517 & 0.009 & 0.100 & 0.513 \\
 & 2000& 0.514 & 0.008 & 0.094 & 0.510 \\
\rowcolor{lightgray}
 & 5000& 0.510 & 0.009 & 0.090 & 0.507 \\
 & 10000& 0.506 & 0.008 & 0.088 & 0.504 \\
\hline
\rowcolor{lightgray}
$Dice\_Gradient$ & 500& 0.517 & 0.009 & 0.097 & 0.513 \\
 & 1000& 0.516 & 0.009 & 0.101 & 0.513 \\
\rowcolor{lightgray}
 & 2000& 0.518 & 0.008 & 0.101 & 0.513 \\
 & 5000& 0.518 & 0.009 & 0.100 & 0.512 \\
\rowcolor{lightgray}
 & 10000& 0.515 & 0.008 & 0.101 & 0.512 \\
\hline
$Dice\_K-DTree$ & 500& 0.517 & 0.011 & 0.110 & 0.517 \\
\rowcolor{lightgray}
 & 1000& 0.513 & 0.009 & 0.104 & 0.512 \\
 & 2000& 0.507 & 0.010 & 0.105 & 0.509 \\
\rowcolor{lightgray}
 & 5000& 0.509 & 0.013 & 0.109 & 0.511 \\
 & 10000& 0.506 & 0.013 & 0.112 & 0.509 \\
\hline
\rowcolor{lightgray}
$SCFE$ & 500& 0.524 & 0.007 & 0.086 & 0.514 \\
 & 1000& 0.524 & 0.008 & 0.109 & 0.517 \\
\rowcolor{lightgray}
 & 2000& 0.520 & 0.008 & 0.093 & 0.510 \\
 & 5000& 0.500 & 0.008 & 0.087 & 0.495 \\
\rowcolor{lightgray}
 & 10000& 0.488 & 0.008 & 0.083 & 0.488 \\
\hline
\end{tabular}
\caption{Adult - Fixed Attack Set Size 1000.}
\label{tab:Adult_attack_1000}
\end{table*}

\begin{table*}[ht]
\centering
\begin{tabular}{llllll}
\hline
\textbf{CF} & \textbf{Synth Size} &  \textbf{ROC AUC} & \textbf{$TPR@FPR.01$} & \textbf{$TPR@FPR.1$} & \textbf{PR AUC} \\
\hline
$NICE$ & 500& 0.626 & 0.013 & 0.201 & 0.609 \\
\rowcolor{lightgray}
 & 1000& 0.618 & 0.012 & 0.184 & 0.601 \\
 & 2000& 0.616 & 0.011 & 0.177 & 0.601 \\
\rowcolor{lightgray}
 & 5000& 0.615 & 0.011 & 0.176 & 0.600 \\
 & 10000& 0.616 & 0.010 & 0.176 & 0.600 \\
\hline
\rowcolor{lightgray}
$Dice\_Gradient$ & 500& 0.619 & 0.017 & 0.179 & 0.594 \\
 & 1000& 0.612 & 0.020 & 0.219 & 0.602 \\
\rowcolor{lightgray}
 & 2000& 0.603 & 0.024 & 0.204 & 0.593 \\
 & 5000& 0.603 & 0.019 & 0.202 & 0.593 \\
\rowcolor{lightgray}
 & 10000& 0.602 & 0.020 & 0.201 & 0.592 \\
\hline
$Dice\_K-DTree$ & 500& 0.609 & 0.009 & 0.200 & 0.611 \\
\rowcolor{lightgray}
 & 1000& 0.620 & 0.008 & 0.188 & 0.611 \\
 & 2000& 0.618 & 0.008 & 0.180 & 0.606 \\
\rowcolor{lightgray}
 & 5000& 0.617 & 0.005 & 0.178 & 0.606 \\
 & 10000& 0.620 & 0.011 & 0.200 & 0.610 \\
\hline
\rowcolor{lightgray}
$SCFE$ & 500& 0.706 & 0.024 & 0.396 & 0.701 \\
 & 1000& 0.697 & 0.031 & 0.357 & 0.689 \\
\rowcolor{lightgray}
 & 2000& 0.694 & 0.026 & 0.336 & 0.684 \\
 & 5000& 0.694 & 0.027 & 0.336 & 0.683 \\
\rowcolor{lightgray}
 & 10000& 0.695 & 0.030 & 0.335 & 0.684 \\
\hline
\end{tabular}
\caption{Compas - Fixed Attack Set Size 500.}
\label{tab:Compas_attack_500}
\end{table*}

\begin{table*}[ht]
\centering
\begin{tabular}{llllll}
\hline
\textbf{CF} & \textbf{Synth Size} &  \textbf{ROC AUC} & \textbf{$\mathbf{TPR@FPR.01}$} & \textbf{$\mathbf{TPR@FPR.1}$} & \textbf{PR AUC} \\
\hline
$NICE$ & 500& 0.500 & 0.012 & 0.105 & 0.506 \\
\rowcolor{lightgray}
 & 1000& 0.511 & 0.019 & 0.123 & 0.516 \\
 & 2000& 0.512 & 0.021 & 0.126 & 0.522 \\
\rowcolor{lightgray}
 & 5000& 0.512 & 0.021 & 0.126 & 0.522 \\
 & 10000& 0.513 & 0.021 & 0.126 & 0.522 \\
\hline
\rowcolor{lightgray}
$Dice\_Gradient$ & 500& 0.487 & 0.009 & 0.091 & 0.492 \\
 & 1000& 0.491 & 0.009 & 0.097 & 0.496 \\
\rowcolor{lightgray}
 & 2000& 0.486 & 0.009 & 0.090 & 0.494 \\
 & 5000& 0.486 & 0.008 & 0.090 & 0.494 \\
\rowcolor{lightgray}
 & 10000& 0.486 & 0.009 & 0.091 & 0.494 \\
\hline
$Dice\_K-DTree$ & 500& 0.519 & 0.050 & 0.147 & 0.542 \\
\rowcolor{lightgray}
 & 1000& 0.536 & 0.055 & 0.170 & 0.557 \\
 & 2000& 0.537 & 0.061 & 0.175 & 0.564 \\
\rowcolor{lightgray}
 & 5000& 0.537 & 0.060 & 0.175 & 0.564 \\
 & 10000& 0.537 & 0.060 & 0.175 & 0.564 \\
\hline
\rowcolor{lightgray}
$SCFE$ & 500& 0.505 & 0.011 & 0.116 & 0.507 \\
 & 1000& 0.502 & 0.010 & 0.102 & 0.503 \\
\rowcolor{lightgray}
 & 2000& 0.496 & 0.009 & 0.097 & 0.498 \\
 & 5000& 0.496 & 0.008 & 0.097 & 0.498 \\
\rowcolor{lightgray}
 & 10000& 0.496 & 0.008 & 0.097 & 0.498 \\
\hline
\end{tabular}
\caption{Heloc - Fixed Attack Set Size 1000.}
\label{tab:Heloc_attack_1000}
\end{table*}

\FloatBarrier

\section{Evaluating the effects of Proximity, Diversity and Actionability on No-box attack performance} \label{app:hyperparameters}
To study in more detail how counterfactual properties affect the attack performance, we generated various versions of the dice attack using different settings for Proximity, Diversity, and Actionability.
For proximity and diversity variation, we implemented 9 different combinations of proximity/diversity weights in the dice explainer setting: \\
\begin{itemize}
    \item $Proximity\_weight \in (.5,1,2)$.
    \item $Diversity\_weight \in (.1,.5,1)$.
\end{itemize}
To address Actionability, we applied the limitation feature\_to\_vary in the dice setting. 
First, we allowed all feature values to change when generating counterfactuals. 
In the second set of experiments, we limited dice to only change numerical features. 
Results of these experiments are presented in tables~\ref{tab:acs_income_dice_variations},~\ref{tab:compas_dice_variations}. 
\begin{table*}[ht]
\centering
\begin{tabular}{llllll}
\hline
\textbf{CF} & \textbf{ROC AUC} & \textbf{$\mathbf{TPR@FPR.01}$} & \textbf{$\mathbf{TPR@FPR.1}$} & \textbf{PR AUC} \\
\hline
dice\_gradient\_pw0.5\_dw0.1\_ftv-all & 0.502 & 0.010 & 0.106 & 0.506 \\
\hline
\rowcolor{lightgray}
dice\_gradient\_pw0.5\_dw0.1\_ftv\_actionable & 0.519 & \textbf{0.029} & 0.098 & 0.518 \\
\hline
dice\_gradient\_pw0.5\_dw0.5\_ftv-all & 0.502 & 0.010 & 0.106 & 0.506 \\
\hline
\rowcolor{lightgray}
dice\_gradient\_pw0.5\_dw0.5\_ftv\_actionable & \textbf{0.520} & \textbf{0.029} & 0.098 & \textbf{0.519} \\
\hline
dice\_gradient\_pw0.5\_dw1\_ftv-all & 0.502 & 0.010 & 0.106 & 0.506 \\
\hline
\rowcolor{lightgray}
dice\_gradient\_pw0.5\_dw1\_ftv\_actionable & \textbf{0.520} & 0.029 & 0.098 & \textbf{0.519} \\
\hline
dice\_gradient\_pw1\_dw0.1\_ftv-all & 0.509 & 0.015 & 0.116 & 0.512 \\
\hline
\rowcolor{lightgray}
dice\_gradient\_pw1\_dw0.1\_ftv\_actionable & 0.506 & 0.011 & 0.105 & 0.506 \\
\hline
dice\_gradient\_pw1\_dw0.5\_ftv-all & 0.509 & 0.015 & 0.116 & 0.512 \\
\hline
\rowcolor{lightgray}
dice\_gradient\_pw1\_dw0.5\_ftv\_actionable & 0.494 & 0.016 & 0.098 & 0.503 \\
\hline
dice\_gradient\_pw1\_dw1\_ftv-all & 0.502 & 0.015 & \textbf{0.116} & 0.508 \\
\hline
\rowcolor{lightgray}
dice\_gradient\_pw1\_dw1\_ftv\_actionable & 0.495 & 0.017 & 0.111 & 0.508 \\
\hline
dice\_gradient\_pw2\_dw0.1\_ftv-all & 0.496 & 0.015 & 0.101 & 0.504 \\
\hline
\rowcolor{lightgray}
dice\_gradient\_pw2\_dw0.1\_ftv\_actionable & 0.496 & 0.010 & 0.098 & 0.502 \\
\hline
dice\_gradient\_pw2\_dw0.5\_ftv-all & 0.496 & 0.012 & 0.105 & 0.504 \\
\hline
\rowcolor{lightgray}
dice\_gradient\_pw2\_dw0.5\_ftv\_actionable & 0.497 & 0.010 & 0.098 & 0.502 \\
\hline
dice\_gradient\_pw2\_dw1\_ftv-all & 0.496 & 0.011 & 0.105 & 0.504 \\
\hline
\rowcolor{lightgray}
dice\_gradient\_pw2\_dw1\_ftv\_actionable & 0.496 & 0.010 & 0.098 & 0.502 \\
\hline
\end{tabular}
\caption{$Acs\_Income$ - evaluation of attack performance across various proximity/diversity/actionability settings for Dice Gradient mechanisms.}
\label{tab:acs_income_dice_variations}
\end{table*}


\begin{table*}[ht]
\centering
\begin{tabular}{llllll}
\hline
\textbf{CF} & \textbf{ROC AUC} & \textbf{$\mathbf{TPR@FPR.01}$} & \textbf{$\mathbf{TPR@FPR.1}$} & \textbf{PR AUC} \\
\hline
dice\_gradient\_pw0.5\_dw0.1\_ftv-all & 0.602 & 0.016 & 0.231 & 0.601 \\
\hline
\rowcolor{lightgray}
dice\_gradient\_pw0.5\_dw0.1\_ftv\_actionable & 0.563 & 0.008 & 0.119 & 0.543 \\
\hline
dice\_gradient\_pw0.5\_dw0.5\_ftv-all & 0.602 & 0.016 & 0.230 & 0.601 \\
\hline
\rowcolor{lightgray}
dice\_gradient\_pw0.5\_dw0.5\_ftv\_actionable & 0.566 & 0.009 & 0.121 & 0.544 \\
\hline
dice\_gradient\_pw0.5\_dw1\_ftv-all & 0.603 & 0.017 & 0.229 & 0.601 \\
\hline
\rowcolor{lightgray}
dice\_gradient\_pw0.5\_dw1\_ftv\_actionable & 0.569 & 0.009 & 0.126 & 0.548 \\
\hline
dice\_gradient\_pw1\_dw0.1\_ftv-all & 0.620 & 0.023 & \textbf{0.283} & 0.626 \\
\hline
\rowcolor{lightgray}
dice\_gradient\_pw1\_dw0.1\_ftv\_actionable & 0.585 & 0.010 & 0.142 & 0.563 \\
\hline
dice\_gradient\_pw1\_dw0.5\_ftv-all & \textbf{0.621} & 0.023 & \textbf{0.283} & \textbf{0.626} \\
\hline
\rowcolor{lightgray}
dice\_gradient\_pw1\_dw0.5\_ftv\_actionable & 0.585 & 0.010 & 0.142 & 0.563 \\
\hline
dice\_gradient\_pw1\_dw1\_ftv-all & 0.620 & \textbf{0.024} & 0.283 & \textbf{0.626} \\
\hline
\rowcolor{lightgray}
dice\_gradient\_pw1\_dw1\_ftv\_actionable & 0.585 & 0.010 & 0.142 & 0.563 \\
\hline
dice\_gradient\_pw2\_dw0.1\_ftv-all & 0.619 & 0.020 & 0.264 & 0.620 \\
\hline
\rowcolor{lightgray}
dice\_gradient\_pw2\_dw0.1\_ftv\_actionable & 0.592 & 0.013 & 0.142 & 0.567 \\
\hline
dice\_gradient\_pw2\_dw0.5\_ftv-all & 0.619 & 0.020 & 0.264 & 0.620 \\
\hline
\rowcolor{lightgray}
dice\_gradient\_pw2\_dw0.5\_ftv\_actionable & 0.592 & 0.013 & 0.142 & 0.567 \\
\hline
dice\_gradient\_pw2\_dw1\_ftv-all & 0.619 & 0.020 & 0.265 & 0.620 \\
\hline
\rowcolor{lightgray}
dice\_gradient\_pw2\_dw1\_ftv\_actionable & 0.592 & 0.012 & 0.141 & 0.567 \\
\hline
\end{tabular}
\caption{$Compas$ - evaluation of attack performance across various proximity/diversity/actionability settings for Dice Gradient mechanisms.}
\label{tab:compas_dice_variations}
\end{table*}

\FloatBarrier

\section{Attack analysis based on distance to the decision boundary} ~\label{app:decision-boundary}
To evaluate the effectiveness of the no-box attack regarding distance to the decision boundary, we performed some extra experiments. 
In this set of experiments, we divided input instances into five bins and generated counterfactuals for each bin. Then, the attack was performed following the setting explained in the main body of the paper on each bin. 
To measure the distance to the decision boundary, we used confidence score as a proxy for distance to the decision boundary~\cite{shokri2021privacy}. 
This way, the higher the confidence score is, the further the instance is from the decision boundary.
Table~\ref{tab:acs_income_confidence_bins} illustrates the results of these experiments for $ACS\_income$ using \texttt{NICE} and $dice\_gradient$ counterfactual generation mechanisms.
These results show that by increasing the confidence score (which means more distant instances from the decision boundary), the attack success rate slightly decreases, which complies with the expectations according to the existing studies~\cite{shokri2021privacy,pawelczyk2023privacy}.

\begin{table*}[ht]
 \centering
 \begin{tabular}{lllllll}
 \hline
 \textbf{CF} & \textbf{Confidence Bin} & \textbf{ROC AUC} & \textbf{$\mathbf{TPR@FPR.01}$} & \textbf{$\mathbf{TPR@FPR.1}$} & \textbf{PR AUC} \\
 \hline
 NICE & 50-60 & 0.509 & 0.006 & 0.102 & 0.506 \\
 \rowcolor{lightgray}
  & 60-70 & \textbf{0.511} & 0.011 & 0.095 & 0.505 \\
  & 70-80 & 0.508 & 0.012 & \textbf{0.111} & 0.512 \\
 \rowcolor{lightgray}
  & 80-90 & 0.510 & 0.016 & 0.111 & \textbf{0.512} \\
  & 90-100 & 0.494 & 0.008 & 0.102 & 0.500 \\
 \hline
\rowcolor{lightgray}
  dice\_gradient & 50-60 & 0.510 & 0.014 & 0.104 & 0.509 \\
 & 60-70 & 0.500 & 0.015 & 0.098 & 0.504 \\
 \rowcolor{lightgray}
 & 70-80 & 0.503 & 0.014 & 0.100 & 0.505 \\
  & 80-90 & 0.503 & \textbf{0.016} & 0.107 & 0.508 \\
 \rowcolor{lightgray}
 & 90-100 & 0.500 & 0.012 & 0.102 & 0.503 \\
 \hline
\end{tabular}
\caption{$Acs\_Income$ - evaluation of attack performance with confidence bins.}
\label{tab:acs_income_confidence_bins}
\end{table*}

\end{document}